\pdfoutput=1

\documentclass[11pt]{article}

\usepackage[]{acl}

\usepackage{times}
\usepackage{latexsym}

\usepackage[T1]{fontenc}

\usepackage[utf8]{inputenc}

\usepackage{microtype}

\usepackage{inconsolata}

\usepackage{hyperref}       
\usepackage{url}            
\usepackage{booktabs}       
\usepackage{amsfonts}       
\usepackage{nicefrac}       

\usepackage{lipsum}		
\usepackage{graphicx}
\usepackage{natbib}
\usepackage{doi}
\usepackage{wrapfig}
\usepackage{booktabs}
\usepackage{array}
\usepackage{subcaption} 
\usepackage{multirow} 

\usepackage[utf8]{inputenc} 
\usepackage[T1]{fontenc}    
\usepackage{hyperref}       
\usepackage{url}            
\usepackage{amsfonts}       
\usepackage{nicefrac}       
\usepackage{microtype}      
\usepackage{doi}
\usepackage{xcolor}         
\definecolor{mypurple}{RGB}{200, 125, 245}
\definecolor{myred}{RGB}{238, 34, 12}
\definecolor{mygrey}{RGB}{146, 146, 146}
\usepackage{tabularx}
\usepackage{makecell}
\usepackage{xspace}
\usepackage{enumitem}
\usepackage{float}
\usepackage{amsmath}
\usepackage{listings}
\usepackage{algorithm}
\usepackage{algorithmic}
\usepackage{mathrsfs}
\usepackage{longtable}
\usepackage{pythonhighlight}

\title{DialogBench: Evaluating LLMs as Human-like Dialogue Systems}

\author{ 
Jiao Ou\textsuperscript{\rm 1}, \textbf{Junda Lu\textsuperscript{\rm 1}}, Che Liu\textsuperscript{\rm 1}, \textbf{Yihong Tang\textsuperscript{\rm 1}}, \\ \textbf{Fuzheng Zhang\textsuperscript{\rm 1}}, \textbf{Di Zhang\textsuperscript{\rm 1}}, \textbf{Kun Gai\textsuperscript{\rm 1}}\\ 
\textsuperscript{\rm 1} Kuaishou\\
{ojiao1111@gmail.com, enbiwudi123@gmail.com}
}

\begin{document}
\maketitle

\begin{abstract}
Large language models (LLMs) have achieved remarkable breakthroughs in new dialogue capabilities by leveraging instruction tuning, which refreshes human impressions of dialogue systems.
The long-standing goal of dialogue systems is to be human-like enough to establish long-term connections with users. 
Therefore, there has been an urgent need to evaluate LLMs as human-like dialogue systems.
In this paper, we propose DialogBench, a dialogue evaluation benchmark that contains $12$ dialogue tasks to probe the capabilities of LLMs as human-like dialogue systems should have.
Specifically, we prompt GPT-4 to generate evaluation instances for each task.
We first design the basic prompt based on widely used design principles and further mitigate the existing biases to generate higher-quality evaluation instances.
Our extensive tests on English and Chinese DialogBench of $26$ LLMs show that instruction tuning improves the human likeness of LLMs to a certain extent, but most LLMs still have much room for improvement as human-like dialogue systems.
Interestingly, results also show that the positioning of assistant AI can make instruction tuning weaken the human emotional perception of LLMs and their mastery of information about human daily life~\footnote{https://github.com/kwai/DialogBench}.

\end{abstract}

\section{Introduction}
Large language models (LLMs)~\citep{bai2022training,du2022glm,sun2023moss,OpenAI2023GPT4TR} have achieved remarkable breakthroughs by leveraging instruction tuning~\citep{wei2021finetuned}, 
especially unlocking new dialogue capabilities.
Such new dialogue capabilities empower humans to naturally interact with LLMs, which has refreshed human’s impression of dialogue systems.
The long-standing goal of dialogue systems requires LLMs to be sufficiently human-like to establish long-term connections with users by satisfying the need for communication, affection and social belonging.
Specifically, human-likeness generally covers the following fine-grained capabilities: correctly understanding the dialogue context, making reasonable use of relevant knowledge, detecting the user's emotions and personality when necessary, and finally generating friendly and reasonable responses that are coherent and consistent with the dialogue context~\citep{huang2020challenges}.
However, the heightened human likeness could not correspond to improved scores on existing LLM benchmarks.

Existing LLM benchmarks are mostly oriented to evaluate the LLMs' abilities for task completion as assistant AI, such as human-knowledge mastery~\citep{zhao2023survey,zeng2023measuring,huang2023c,cobbe2021training} or instruction following~\citep{mishra2022cross,zheng2023judging}.
However, these benchmarks do not focus on whether LLMs as dialogue systems are sufficiently human-like to establish long-term connections with users.
Therefore, an in-depth evaluation benchmark of those abilities related to human likeness is essential for identifying the strengths and limitations of LLMs as multi-turn dialogue systems.

The most ideal approach is to collect corresponding high-quality dialogues from real humans.
However, most real-human dialogues, whether from social networks or open datasets, are likely to have been leaked during the pre-training of LLMs.
To prevent the issue of ``data leakage'', the evaluation benchmark must contain new evaluation instances and be updated frequently.
Due to the difficulty of human writing, it is necessary to construct new human-human dialogues as evaluation instances automatically.
Inspired by~\citet{moller2023prompt} and~\citet{whitehouse-etal-2023-llm}, we explore the use of GPT-4 as a surrogate for humans to generate massive evaluation instances.

In this paper, we propose a novel \textbf{Dialog}ue Evaluation \textbf{Bench}mark with GPT-4 as Data Generator, DialogBench for short.
Since dialogues generated without restrictions may not involve commonsense use or emotional expression, we generate corresponding evaluation instances for different fine-grained capabilities.
To evaluate comprehensive abilities, we select 12 dialogue tasks. Each task requires LLMs to possess at least one ability to perform it well.
For each task, we prompt GPT-4 to generate evaluation instances.
Specifically, we first design the basic prompt based on widely-used design principles and further mitigate the existing biases to generate most of the available evaluation instances.
Afterward, we filter out detrimental evaluation instances via a filter mechanism.
Consequently, we construct English and Chinese dialogue evaluation benchmarks towards human likeness.

We conduct a comprehensive evaluation of $26$ LLMs using DialogBench, including pre-trained and supervised instruction-tuning models. 
Experimental results reveal that instruction tuning can improve the human likeness of LLMs.
For supervised instruction-tuning models, top-tier models can handle a wide array of dialogue tasks, indicating the potential for developing LLMs into human-like dialogue systems.
However, we observe significant performance gaps between top-tier models and other LLMs, which suggests that their performance lags considerably.
In addition, LLMs generally perform better at correctly understanding context but are relatively poor at perceiving emotions and personality.
Current LLMs also do not understand much about daily human life.
This underscores the necessity for more efforts to enhance the abilities related to the human likeness of most LLMs.

Our contributions are summarized as follows: 
(1) We present DialogBench, a comprehensive benchmark to standardize the evaluation of LLMs as human-like dialogue systems. 
(2) We perform a thorough evaluation of $26$ different LLMs using DialogBench, uncovering a significant performance evaluation under diverse dialogue tasks. It illuminates the top-tier LLM in human likeness and highlights dimensions for improvement.
\begin{figure*}
\centering
\includegraphics[width=0.95\textwidth,keepaspectratio]{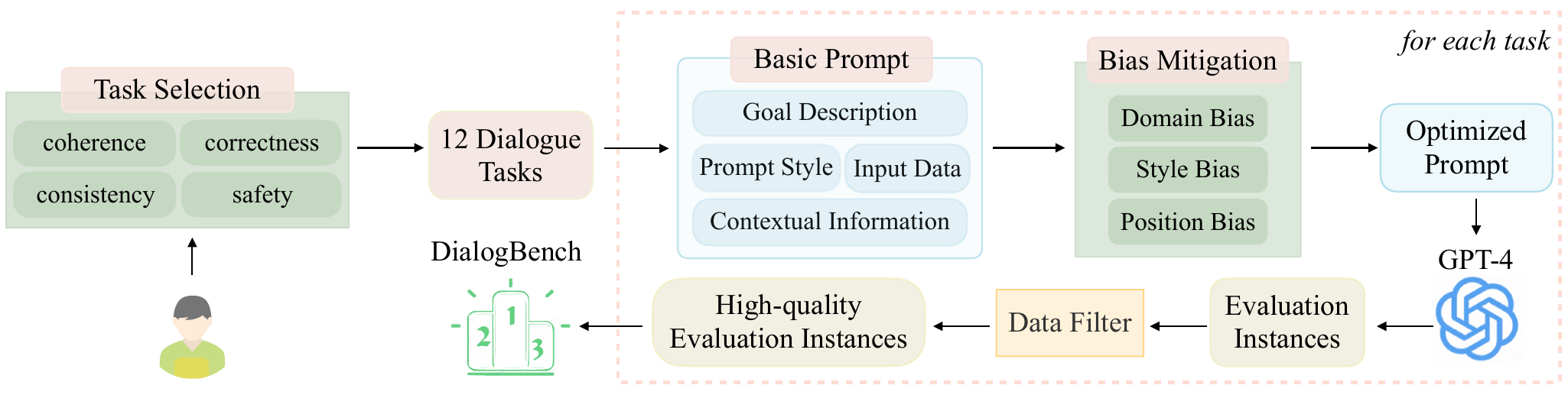}
\caption{The overall architecture of DialogBench construction.}
\label{app:overall_architecture}
\end{figure*}

\section{Related Work}
\paragraph{Evaluation of LLMs.}
To better understand LLM's strengths and limitations, many benchmarks are proposed to evaluate broad capabilities.
These benchmarks mainly evaluate the LLMs' ability to complete tasks as an assistant AI and can be divided into the following categories~\citep{zhao2023survey}.
Comprehensive-evaluation benchmarks~\citep{liang2022holistic,srivastava2023beyond,li2023cmmlu,choi-etal-2023-llms} are applied to holistically evaluate LLMs on multiple NLP tasks.
Human-centric benchmarks~\citep{zeng2023measuring,zhong2023agieval,huang2023c,xu2023superclue,clarkthink} primarily focus on evaluation in human-centric scenarios by collecting qualification exams.
In addition, special-ability benchmarks~\citep{ahn2022can,liu2023agentbench,li-etal-2023-api,babe2023studenteval,chalamalasetti-etal-2023-clembench} place more emphasis on advanced abilities.
Despite the emergence of various benchmarks, no benchmark comprehensively evaluates LLMs as human-like dialogue systems.

\paragraph{Dialogue Benchmarks.}

There are several benchmarks for evaluating dialogue capabilities~\citep{reddy-etal-2019-coqa,mehri2020dialoglue,gupta-etal-2022-instructdial}.
These benchmarks can be used to evaluate language models that have been fine-tuned on the corresponding training sets but cannot directly evaluate instruction-following LLMs.
In addition, these previous benchmarks may have been leaked during the pre-training of LLMs.
In contrast, DialogBench contains new evaluation instances with natural language, which can be directly used to evaluate instruction-following LLMs and avoid data leakage.
\citet{zheng2023judging} evaluates LLMs’ multi-turn instruction-following abilities, which focuses on assessing its alignment with human preference, rather than LLMs as human-like dialogue systems.
Recent researchers ~\cite{zhao2023chatgpt,wang2023chatgpt,rao2023can,ji2023chatgpt,wang-etal-2023-cue} also focus on human-like characters of GPT-4 or ChatGPT.
However, our work holistically evaluates capabilities related to human likeness.

\paragraph{LLMs for Data Generation.}

Many recent researches~\citep{whitehouse-etal-2023-llm,yu2023large,tang2023does,Xu2023BaizeAO,whitehouse-etal-2023-llm} also leverage GPT-4 for data generation, mainly using several training instances as few-shot examples to prompt the generation of more training instances.
In contrast, our work leverages GPT-4 to generate new evaluation instances for constructing benchmarks without few-shot examples.

\section{DialogBench}
\label{app:bench}

\begin{figure}
\centering
    \includegraphics[width=0.85\linewidth,keepaspectratio]{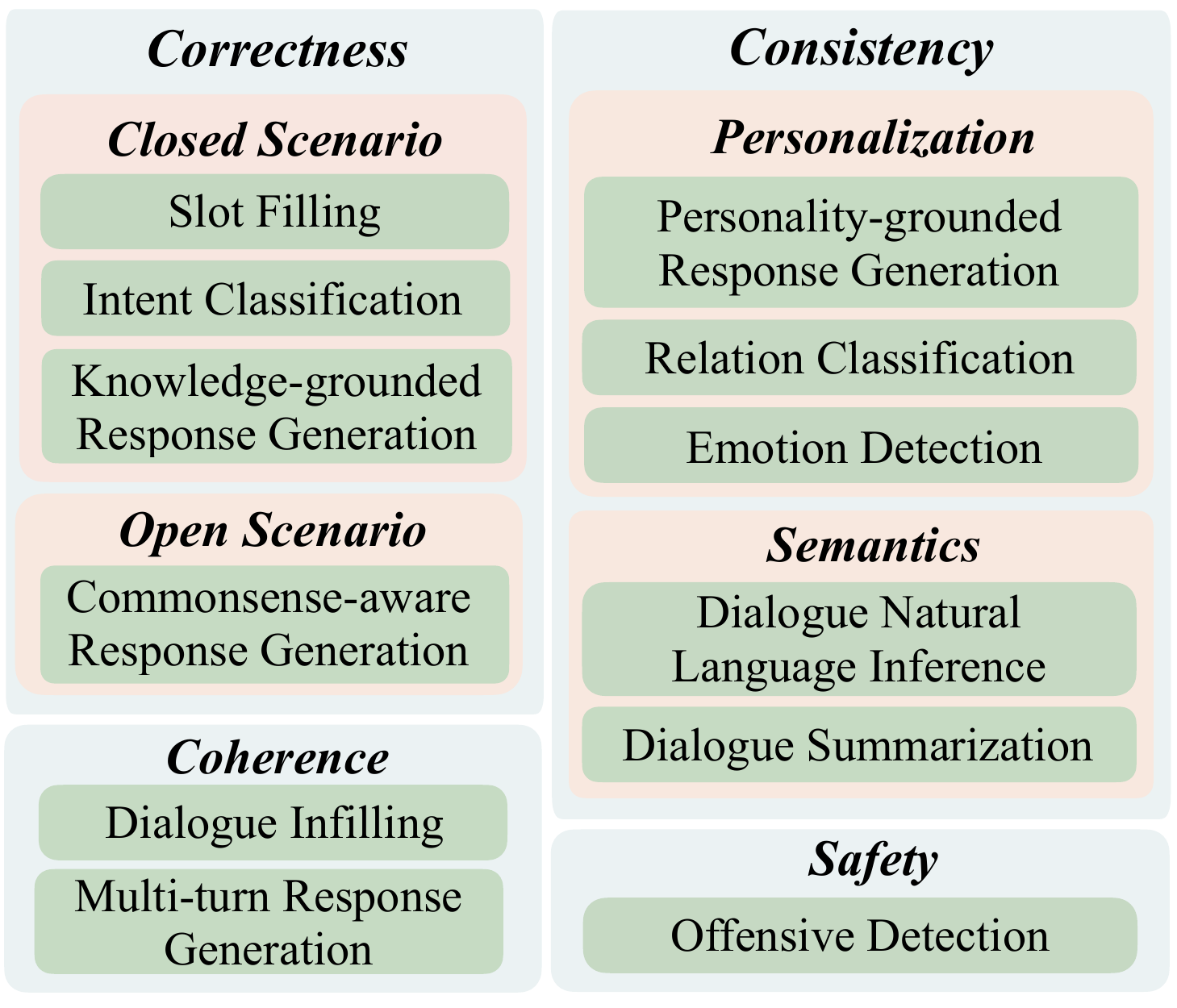}
\caption{Task selection in DialogBench.}

\label{fig:task_selection}
\end{figure}
In this section, our goal is to generate evaluation instances using GPT-4.
To this end, in section~\ref{app:task_selection}, we describe the selection of dialogue tasks.
In section~\ref{app:metric}, we describe how to determine the question type of evaluation instances, like generation questions or multi-choice questions, to effectively reflect the quality of LLMs as human-like dialogue systems.
In section~\ref{app:prompt}, we design the basic prompt as the input of GPT-4.
In section~\ref{app:quality}, we describe the biases of the basic prompt and the corresponding solutions, along with introducing a filter mechanism to pick out high-quality data.
The overall architecture of DialogBench construction is shown in Figure~\ref{app:overall_architecture}. 

\subsection{Task Selection}
\label{app:task_selection}
To confirm what capabilities LLMs need to have to be like a human, we refer to the main dimensions that are concerned when evaluating human likeness of open-domain dialogue systems, including coherence, consistency, diversity, and fluency~\citep{mehri-eskenazi-2020-usr}.
Considering that LLMs have made great progress in diversity and fluency, along with having more requirements in correctness and safety~\citep{yuan2023gpt,cheng2023gpt}, we refine the evaluation dimensions, including \emph{coherence}, \emph{consistency}, \emph{correctness}, and \emph{safety}.
Consequently, we apply each evaluation dimension as a guide and select tasks that focus on the corresponding evaluation dimension. Accordingly, those abilities can be reflected by the quality of task completion.
Specifically, we elaborately tease out $12$ dialogue tasks. The detailed selection process and task definitions are presented in Appendix~\ref{appendix:task_selection}. The overall selection results are shown in Figure~\ref{fig:task_selection}.

\subsection{Question Setting}
\label{app:metric}

\begin{figure*}
\centering
\includegraphics[width=0.99\textwidth,keepaspectratio]{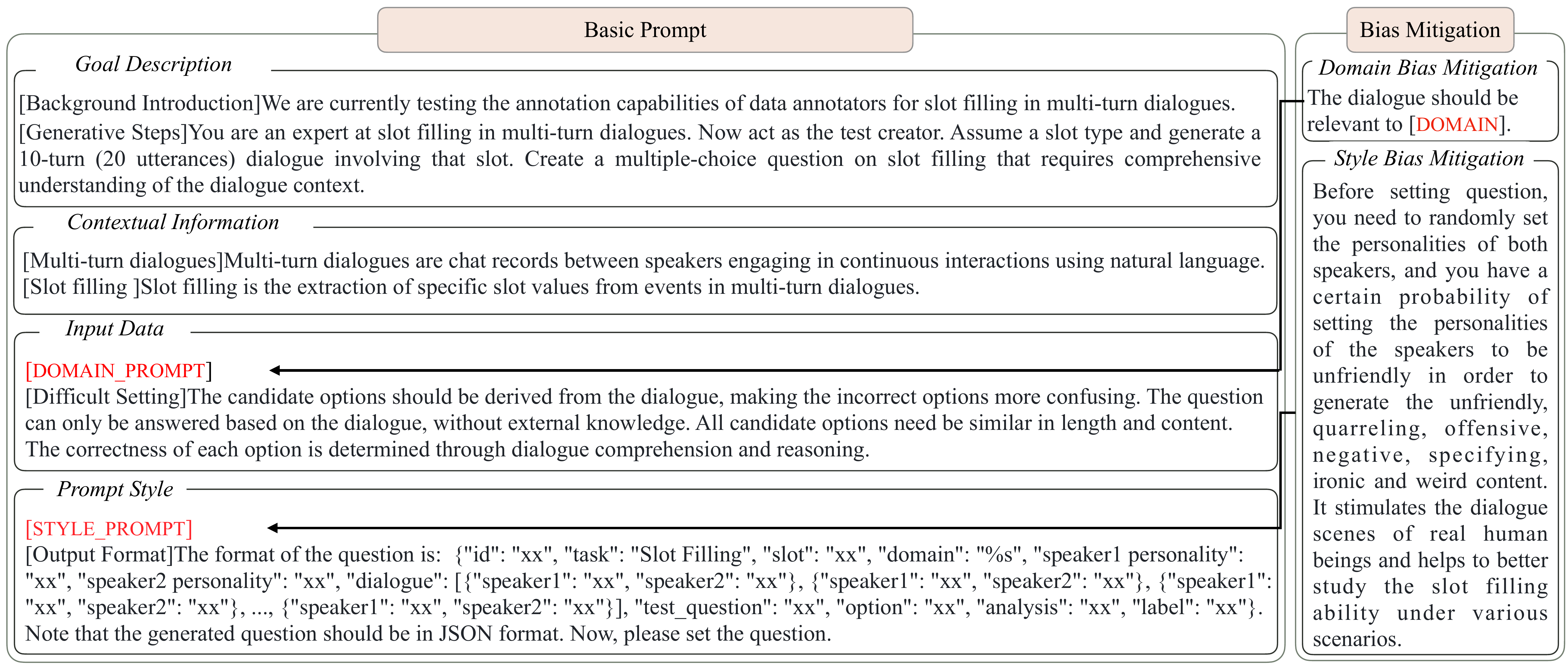}
\caption{The basic prompt on the left and the external description for mitigating bias on the right. We take \emph{slot filling} as an example. \textcolor{red}{DOMAIN} is the placeholder of the given domain. [\textcolor{red}{DOMAIN\_PROMPT}] and [\textcolor{red}{STYLE\_PROMPT}] are the positions of the corresponding description to add.
}
\label{fig:prompt_formatting}
\end{figure*}

\begin{figure}[t!]

\includegraphics[width=0.9\linewidth,keepaspectratio]{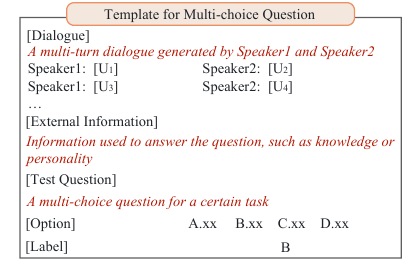}
\caption{The template for multi-choice questions in DialogBench.The \textcolor{myred}{red} text is the explanation.}
\label{fig:question_setting}
\end{figure}

\begin{figure}[t!]
\centering
    \includegraphics[width=0.9\linewidth,keepaspectratio]{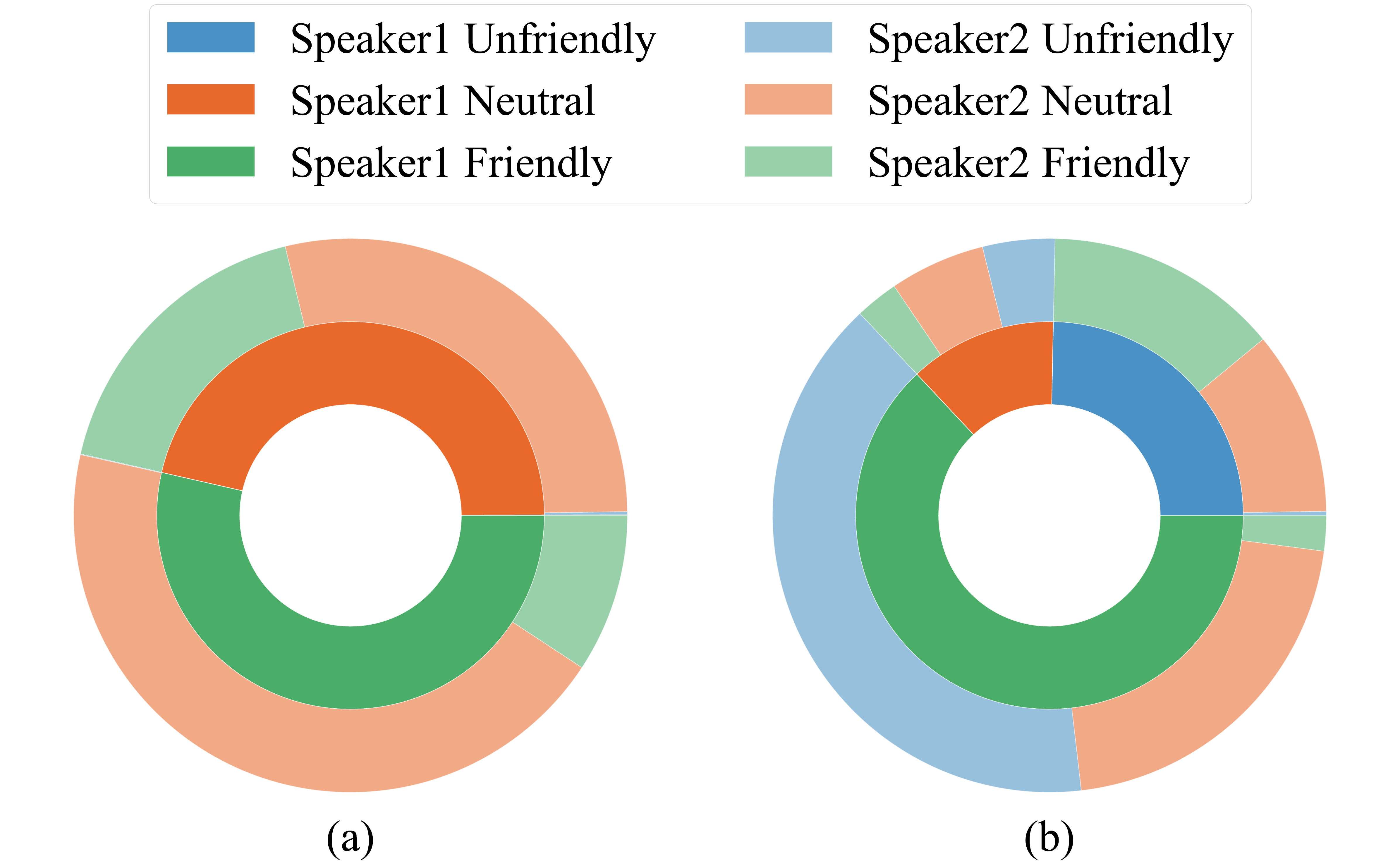}
\caption{(a) The proportion of each dialogue style on both speakers in all generated dialogues via the basic prompt; (b) the proportion via the optimized prompt. 
The inner ring is the proportion of speaker1's style, while the outer ring is the proportion of the corresponding speaker2's style given speaker1's style.
}
\label{fig:style_bias}
\end{figure}

\begin{figure*}
\centering
\includegraphics[width=0.95\textwidth,keepaspectratio]{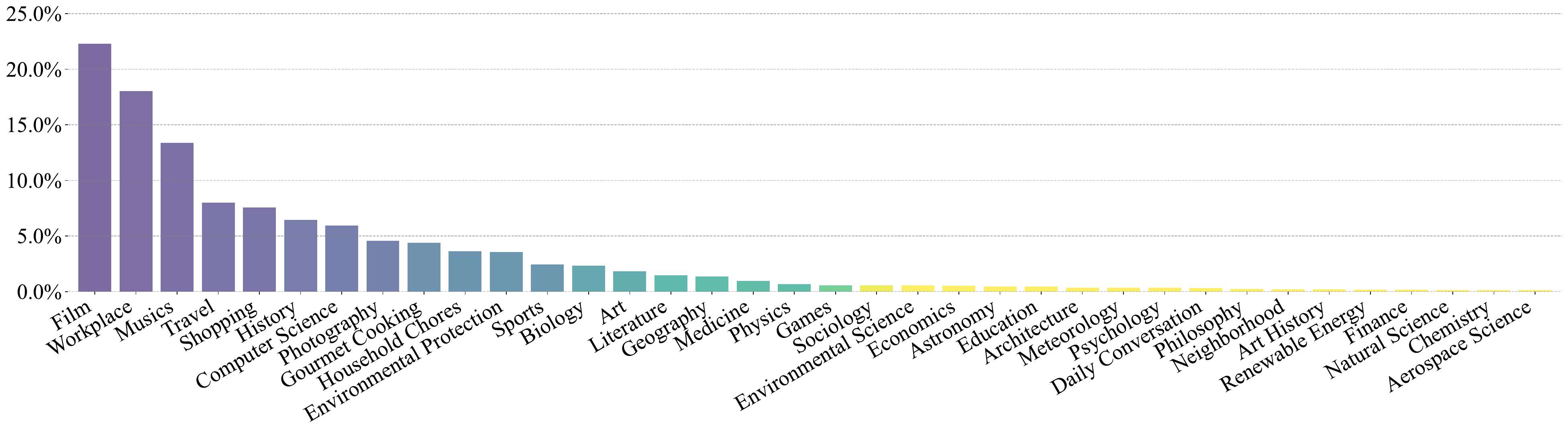}
\caption{The proportion of generated evaluation instances based on the basic prompt in each domain. 
}
\label{fig:damain_bias}
\end{figure*}

\begin{figure*}[t!]
\centering
\includegraphics[width=0.9\textwidth,keepaspectratio]{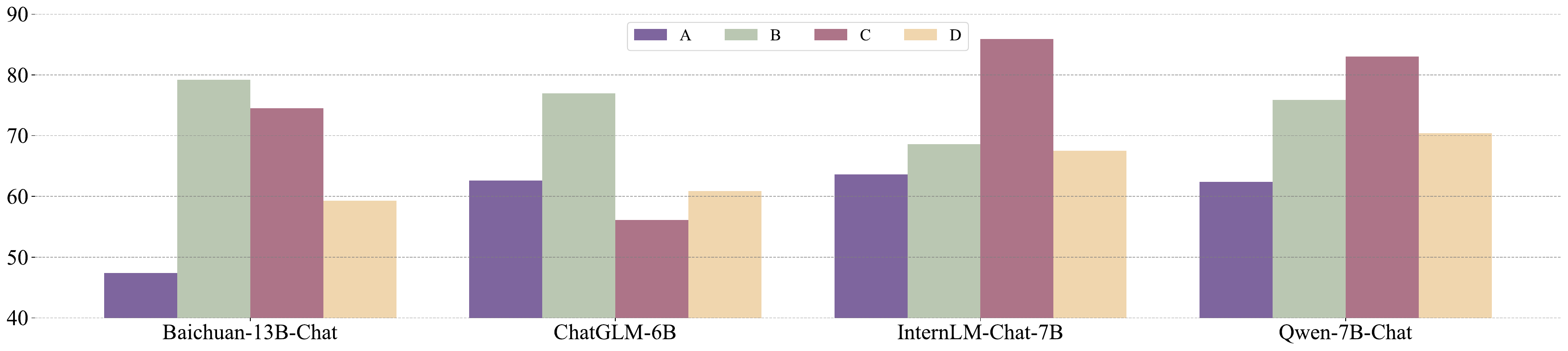}
\caption{The accuracy of LLMs when placing the correct answers of all evaluation instances in specific positions.}
\label{fig:model_position}
\end{figure*}
The selected tasks not only include understanding tasks but also generation tasks, and their corresponding evaluation metrics are different.
To unify evaluation,  we follow most existing benchmarks~\citep{li2023cmmlu,hendrycks2021measuring,huang2023c} to adopt multi-choice questions and use accuracy as the evaluation metric. 
Consequently, an evaluation instance requires LLMs to select the correct answer from candidate options based on the given multi-turn dialogue context for the given test question relevant to the specific task.
The question templates are shown in Figure~\ref{fig:question_setting}.

\subsection{Prompt Formatting}
\label{app:prompt}
A well-designed prompt helps generate high-quality evaluation instances.
We create prompts according to the prompt design proposed by~\citet{zhao2023survey}, which summarizes four key ingredients of prompts and several basic design principles.
Specifically, we take \emph{slot filling} as an example to describe the prompt creation.
We first clarify the core content based on these key ingredients and then integrate them into an effective prompt based on the design principles.
The detailed creation is described in Appendix~\ref{appendix:prompt_formatting}.
The final prompt is the exact string that concatenates each content of the four ingredients, shown in Figure~\ref{fig:prompt_formatting}.

\subsection{Quality Control}
\label{app:quality}
We observe several biases 
and low-quality instances in generated evaluation instances.
Next, we present the corresponding solutions for mitigating biases and filter mechanisms.
The optimized prompts for all tasks are shown in Table~\ref{tab:ic_prompt}-\ref{tab:ds_prompt}.

\paragraph{Bias Mitigation.}

For domain bias, we first count the number of each domain covered by all evaluation instances. 
Specifically, we employ GPT-4 to detect the domain that the given dialogue context is about.
Their statistics are shown in Figure~\ref{fig:damain_bias}, which show that GPT-4 tends to generate several common domains, leading to a long-tail distribution.
This may cause two issues: (1) The imbalanced numbers in each domain will cause the overall results to be biased; (2) The results in domains with insufficient data are not accurate enough.
Thus, it is necessary to balance the amount of instances in each domain and ensure that each domain has enough instances.
By observing the domains involved in human dialogues, we manually designate $20$ domains, mainly involved in two major categories: daily life and professional knowledge.
Specifically, we externally introduce the domain into the ``input data'' of the basic prompt, shown in the right of Figure~\ref{fig:prompt_formatting}.
The domain information is shown in Appendix~\ref{app:domain_bias}. 

For style bias, we observe that the generated dialogues have almost no unfriendly dialogue style.
We roughly divide dialogue styles into friendly, neutral, and unfriendly, and further calculate the proportion of each style on both speakers of all dialogues.
The results are shown in Figure~\ref{fig:style_bias}(a), which shows that GPT-4 hardly generates unfriendly dialogues via the basic prompt.
However, there are quite a few unfriendly dialogues in the real human world. 
Unfriendly communication would greatly increase the difficulty of interaction, and evaluating LLMs in unfriendly scenarios can reflect the true level of LLMs as human-like dialogue systems.
Therefore, we induce GPT-4 to generate a certain proportion of unfriendly dialogues by optimizing the basic prompt.
Since the dialogue style is related to the personalities of both speakers, we require GPT-4 to randomly set their personalities before generating the dialogue.
Inspired from~\citet{moller2023prompt}, we introduce the external information into the basic prompt, displayed in the right of Figure~\ref{fig:prompt_formatting}.
Similarly, We calculate the proportion after mitigating style bias.
The result in Table~\ref{fig:style_bias}(b) indicates that GPT-4 with the optimized prompt can generate a certain percentage of unfriendly dialogues.

For position bias of correct answers,  GPT-4 does not guarantee that the correct answers in all generated evaluation questions are evenly distributed among the candidate options.
Furthermore, we observe that several LLMs have their selection preferences, shown in Figure~\ref{fig:model_position}.
Specifically, we calculate the accuracy of these LLMs when placing the correct answers in different positions on the whole evaluation set.
Therefore, the accuracy of LLMs may be inaccurate when we apply the evaluation instances that GPT-4 generates without correction.
To mitigate position bias, we assign the position of the correct answer among candidate options randomly~\citep{zheng2023judging}.
It can be effective at a large scale with the correct expectations. 

\begin{table}[t!]
\centering
\scriptsize
\resizebox{\linewidth}{!}{
\begin{tabular}{@{}lccc@{}}
\toprule
\textbf{Task} & \textbf{Abbr.} & \textbf{\#Turn} &\textbf{\#Num} \\
\midrule
Knowledge-grounded Response Generation & KRG & 7.41 & 784 \\ Intent Classification & IC & 7.72 & 931 \\
Slot Filling & SF & 7.49 & 879 \\ Emotion Detection & ED & 7.09 & 823 \\
Personality-grounded Response Generation & PRG & 7.16 & 832 \\ Multi-turn Response Generation & MRG & 7.66 & 800 \\
Dialogue Summarization & DS & 9.11 & 738 \\ Commonsense-aware Response Generation & CRG & 7.14 & 709 \\
Dialogue Infilling & DI & 7.68 & 776 \\ Offensive Detection & OD & 8.25 & 802 \\
Dialogue Natural Language Inference & NLI & 6.39 & 882 \\ Relation Classification & RC & 8.56 & 855 \\
\bottomrule
\end{tabular}}

\caption{Statistics of 12 dialogue tasks. ``Abbr.'' denotes the abbreviation. ``\#Turn'' denotes the average dialogue turns. ``\#Num'' denotes the number of instances.}
\label{tab:dataset_stats}
\end{table}

\paragraph{Data Filter.}
The generated evaluation set inevitably contains low-quality instances.
Inspired by~\citet{zhou2022large}, we propose to adopt GPT-4 to filter out low-quality instances. 
We prompt GPT-4 to check whether the multiple-choice questions are correct.
The prompt is displayed in Table~\ref{tab:filter_prompt}.
We further retain only those evaluation instances that GPT-4 considers correct.
It is mainly based on two assumptions: (1) GPT-4 can serve as a surrogate for humans~\citep{zheng2023judging}; (2) a correct instance generated by GPT-4 should be answered correctly by itself.
Through statistics, the average filtering ratio on the whole evaluation set is $10.08\%$.

\section{Experimental Setup}
\paragraph{Dataset Statistics.}
We report the statistics of DialogBench in Table~\ref{tab:dataset_stats}. For simplicity, in the following part,  we use the abbreviation of each task. 

\begin{table*}[t!]
\centering
\resizebox{\textwidth}{!}{
\begin{tabular}{@{}llccccccccccccc@{}}
    \toprule
    
    \multirow{2}{*}{\textbf{Type}}  & \multirow{2}{*}{\textbf{Model}} & \multicolumn{4}{c}{\textbf{Correctness}} & \multicolumn{2}{c}{\textbf{Coherence}} & \multicolumn{5}{c}{\textbf{Consistency}} & \multicolumn{1}{c}{\textbf{Safety}}\\ 
    \cmidrule(lr){3-6} \cmidrule(lr){7-8} \cmidrule(lr){9-13} \cmidrule(lr){14-14}

& & \textbf{SF} & \textbf{IC} & \textbf{KRG} & \textbf{CRG} & \textbf{DI} & \textbf{MRG} & \textbf{PRG} & \textbf{RC} & \textbf{ED} & \textbf{NLI} & \textbf{DS} & \textbf{OD} & \textbf{Overall} \\

\midrule

Human & & 98.00 & 96.00 & 92.00 & 92.00 & 90.00 & 96.00 & 90.00 & 96.00 & 92.00 & 86.00 & 96.00 & 86.00 & 92.50 \\ \midrule \midrule

\multirow{15}{*}{Pre-trained}

& LLaMA2-70B & \textbf{84.94} & \underline{65.88} & \textbf{66.25} & \textbf{62.48} & \underline{44.58} & 51.17 & \textbf{30.43} & \textbf{58.62} & 57.47 & \textbf{67.94} & \textbf{77.24} & 46.02 & \textbf{59.42} \\ 
& LLaMA-65B & \underline{84.83} & 63.65 & 62.40 & 54.90 & 43.19 & 46.17 & 21.45 & 47.36 & \underline{59.20} & 41.63 & 70.76 & \underline{47.50} & \underline{53.59} \\ 
& Baichuan2-13B & 79.31 & 58.95 & 59.50 & 53.73 & 43.34 & 48.50 & \underline{24.93} & 44.60 & \textbf{70.00} & \underline{48.09} & 66.90 & 28.18 & 52.17 \\ 
& Qwen-7B & 69.93 & 59.17 & 63.64 & 56.08 & 42.41 & \underline{51.61} & 20.58 & \underline{52.41} & 56.67 & 45.10 & 63.45 & 44.32 & 52.11 \\ 

& Mistral-7B & 83.56 & \textbf{66.33} & \underline{63.77} & \underline{60.21} & 43.18 & \textbf{53.16} & 18.99 & 18.84 & 57.86 & 45.33 & \underline{76.13} & 35.90 & 51.94 \\

& InternLM-7B & 78.74 & 58.50 & 58.95 & 53.73 & 40.09 & 48.28 & 21.45 & 48.05 & 58.13 & 37.44 & 67.86 & \textbf{49.66} & 51.74 \\ 
& LLaMA2-13B & 81.42 & 60.74 & 60.74 & 57.39 & 43.03 & 47.72 & 24.64 & 30.48 & 57.47 & 42.58 & 71.31 & 41.02 & 51.55 \\ 
& Baichuan-13B & 79.54 & 61.07 & 60.74 & 52.94 & 42.72 & 49.61 & 24.35 & 41.26 & 50.67 & 46.65 & 68.14 & 31.02 & 50.73 \\ 
& LLaMA-7B & 73.45 & 55.70 & 57.44 & 52.68 & 42.72 & 46.50 & 20.29 & 44.83 & 57.20 & 46.17 & 65.10 & 42.27 & 50.36 \\ 
& LLaMA-13B & 76.32 & 59.40 & 58.68 & 54.38 & 40.40 & 39.39 & 19.71 & 47.13 & 59.07 & 40.91 & 65.14 & 41.36 & 50.16 \\ 
& Chinese LLaMA2-13B & 79.43 & 59.84 & 51.71 & 58.43 & \textbf{45.67} & 50.39 & 10.10 & 31.72 & 51.53 & 46.17 & 69.66 & 43.64 & 49.86 \\ 
& Falcon-7B & 75.63 & 57.72 & 54.82 & 47.45 & 39.94 & 42.18 & 15.94 & 40.92 & 55.87 & 40.43 & 62.48 & 37.84 & 47.60 \\ 
& MOSS-Moon-003-Base & 57.93 & 51.01 & 56.06 & 45.88 & 41.02 & 44.73 & 11.88 & 36.55 & 47.33 & 40.43 & 52.97 & 35.57 & 43.45 \\ \midrule
& Avg. & 77.31 & 59.84 & 59.59 & 54.64 & 42.48 & 47.65 & 20.36 & 41.75 & 56.81 & 45.30 & 67.47 & 40.33 & 51.13 \\

\midrule 
\midrule
    
\multirow{15}{*}{\shortstack[l]{Supervised \\ Instruction-tuning}}

        & GPT-4 & \textbf{96.09} & \textbf{93.96} & \textbf{90.01} & \textbf{89.14} & \textbf{85.45} & \textbf{79.00} & \textbf{76.81} & \textbf{88.74} & \textbf{73.87} & \textbf{82.78} & \textbf{92.41} & \textbf{84.47} & \textbf{86.06} \\ 
        & ChatGPT & \underline{89.43} & \underline{83.89} & \underline{83.88} & \underline{84.55} & \underline{75.35} & 75.22 & \underline{62.83} & \underline{83.91} & \underline{68.53} & \underline{74.04} & \underline{86.62} & \underline{68.75} & \underline{78.08} \\ 
        & Baichuan2-13B-Chat & 84.37 & 81.43 & 79.06 & 79.08 & 57.43 & 76.14 & 54.99 & 79.47 & 54.80 & 55.02 & 81.66 & 42.73 & 68.85 \\ 
        & InternLM-Chat-7B & 80.23 & 80.43 & 82.37 & 78.56 & 65.02 & 77.14 & 47.54 & 60.47 & 46.40 & 65.07 & 75.03 & 64.43 & 68.56 \\ 
        & Qwen-7B-Chat & 84.48 & 79.75 & 80.85 & 79.08 & 65.48 & \underline{77.69} & 39.78 & 59.93 & 20.27 & 58.73 & 81.10 & 58.18 & 65.44 \\ 

& Mistral-7B-Instruct & 64.36 & 70.02 & 79.88 & 79.05 & 67.49 & 70.14 & 47.47 & 51.88 & 47.73 & 56.69 & 81.93 & 56.13 & 64.40 \\
        
        & ChatGLM2-6B & 72.64 & 73.94 & 78.10 & 69.02 & 62.69 & 66.81 & 44.06 & 71.49 & 47.87 & 53.11 & 59.45 & 50.23 & 62.45 \\ 
        & Baichuan-13B-Chat  & 74.37 & 71.48 & 73.42 & 70.20 & 50.93 & 72.48 & 45.22 & 72.64 & 49.07 & 39.71 & 68.14 & 50.23 & 61.49 \\ 
        & LLaMA2-7B-Chat & 62.86 & 71.81 & 72.04 & 66.54 & 53.72 & 56.38 & 44.35 & 73.33 & 46.00 & 48.68 & 73.93 & 54.20 & 60.32 \\ 
        & Vicuna-13B & 74.37 & 62.53 & 75.90 & 66.27 & 55.73 & 53.94 & 26.09 & 71.49 & 43.20 & 42.94 & 62.07 & 51.25 & 57.15 \\ 
        & Chinese Alpaca2-13B & 75.52 & 70.36 & 64.19 & 37.78 & 56.19 & 46.50 & 38.26 & 62.76 & 50.27 & 39.47 & 74.21 & 36.70 & 54.35 \\ 
        & MOSS-Moon-003-SFT & 40.00 & 47.20 & 58.82 & 45.10 & 41.33 & 52.83 & 24.06 & 53.79 & 22.93 & 38.52 & 49.66 & 50.57 & 43.73 \\ 
        & Xwin-LM-7B & 48.39 & 52.24 & 46.01 & 42.48 & 33.44 & 37.74 & 26.67 & 56.32 & 22.00 & 30.26 & 52.69 & 31.70 & 40.00 \\ 
        \midrule
        & Avg. & 72.85 & 72.23 & 74.19 & 68.22 & 59.25 & 64.77 & 44.47 & 68.17 & 45.61 & 52.69 & 72.22 & 53.81 & 62.38 \\

\bottomrule
    
\end{tabular}}
\caption{Accuracy on Engilsh DialogBench. 
\textbf{Bold} and \underline{underlined} indicate the best results and second-best results.
}
\label{tab:main_result_en}
\end{table*}

\paragraph{LLMs to Evaluate.}
As a systematic attempt to benchmark existing LLMs (Table~\ref{tab:model_details} in Appendix~\ref{appendix:llm_to_evaluate}) on DialogBench, we include in total $26$ models for evaluation, which could be classified into two categories:
(1) \textbf{Pre-trained LLMs:} which mostly come from the LLaMA model variants or are trained from scratch by academia and companies. All pre-trained LLMs are open-sourced LLMs. 
(2) \textbf{Supervised instruction-tuning LLMs:} which mostly release from the academia and companies. Except for GPT-4 and ChatGPT, the remaining are open-sourced LLMs. 
In addition, we test the \textbf{human level} in these dialogue tasks. 
Specifically, we randomly choose $50$ evaluation instances for each task and then employ $3$ experts to do these questions. Finally, a question is considered correct if at least $2$ experts answer it correctly.
These results can reveal not only the quality of DialogBench but also the human level of this benchmark.

\begin{table*}[t!]
\centering
\resizebox{\textwidth}{!}{
\begin{tabular}{@{}llccccccccccccc@{}}
    \toprule
    \multirow{2}{*}{\textbf{Type}}  & \multirow{2}{*}{\textbf{Model}} & \multicolumn{4}{c}{\textbf{Correctness}} & \multicolumn{2}{c}{\textbf{Coherence}} & \multicolumn{5}{c}{\textbf{Consistency}} & \multicolumn{1}{c}{\textbf{Safety}}\\ 
    \cmidrule(lr){3-6} \cmidrule(lr){7-8} \cmidrule(lr){9-13} \cmidrule(lr){14-14}

& & \textbf{SF} & \textbf{IC} & \textbf{KRG} & \textbf{CRG} & \textbf{DI} & \textbf{MRG} & \textbf{PRG} & \textbf{RC} & \textbf{ED} & \textbf{NLI} & \textbf{DS} & \textbf{OD} & \textbf{Overall} \\  \midrule

        Human & & 96.00 & 96.00 & 96.00 & 94.00 & 90.00 & 94.00 & 96.00 & 96.00 & 94.00 & 86.00 & 98.00 & 84.00 & 93.33 \\ \midrule \midrule
        \multirow{15}{*}{Pre-trained} & Baichuan2-13B & 78.81 & 55.37 & \underline{63.18} & 54.71 & \underline{46.12} & \underline{52.63} & 26.98 & \underline{46.87} & 67.21 & 39.10 & \underline{66.94} & 55.23 & \textbf{54.43} \\ 
        &Qwen-7B & \underline{80.91} & \textbf{61.79} & 63.05 & \textbf{60.57} & 42.52 & \textbf{54.58} & 27.22 & \textbf{56.89} & 59.73 & 22.06 & \textbf{69.52} & 44.77 & \underline{53.63} \\ 
        &InternLM-7B & 75.67 & 56.48 & 61.72 & 55.43 & 44.04 & 47.71 & 26.38 & 45.74 & \textbf{69.93} & \textbf{45.11} & 65.44 & 46.75 & 53.37 \\ 
        &LLaMA2-70B & \textbf{81.84} & \underline{59.56} & \textbf{67.93} & \underline{56.78} & \textbf{47.56} & 43.19 & 27.32 & 24.34 & \underline{69.80} & 39.96 & 44.56 & 56.57 & 51.62 \\ 
        & Mistral-7B & 76.71 & 55.92 & 61.98 & 53.42 & 44.87 & 49.57 & 27.33 & 19.92 & 57.16 & 38.09 & 66.12 & \underline{59.32} & 50.87 \\
        &Baichuan-13B & 75.79 & 54.49 & 60.53 & 54.29 & 44.32 & 47.46 & 24.94 & 39.10 & 39.59 & 34.84 & 65.85 & \textbf{62.57} & 50.31 \\ 
        &LLaMA2-13B & 74.18 & 53.06 & 61.73 & 51.20 & 43.04 & 45.74 & \underline{28.61} & 20.82 & 55.64 & 35.26 & 59.35 & 56.98 & 48.80 \\ 
        &Moss-Moon-003-Base & 61.82 & 48.06 & 59.87 & 52.43 & 41.41 & 47.20 & 25.90 & 32.96 & 60.54 & 38.85 & 60.27 & 54.80 & 48.68 \\ 
        &Chinese LLaMA2-13B & 72.29 & 55.59 & 61.72 & 53.71 & 43.07 & 47.12 & 25.18 & 33.71 & 55.51 & 22.18 & 65.44 & 44.77 & 48.36 \\ 
        &LLaMA-65B & 75.49 & 55.73 & 62.79 & 50.34 & 43.26 & 42.95 & 21.95 & 15.19 & 68.32 & \underline{41.34} & 39.10 & 56.19 & 47.72 \\ 
        &LLaMA-13B & 62.75 & 51.50 & 58.01 & 42.71 & 44.60 & 44.83 & \textbf{28.90} & 13.16 & 57.82 & 39.22 & 56.05 & 55.23 & 46.23 \\ 
        &LLaMA-7B & 62.65 & 49.39 & 58.81 & 42.29 & 45.15 & 44.58 & 27.94 & 13.41 & 35.10 & 40.23 & 55.10 & 55.37 & 44.17 \\ 
        &Falcon-7B & 65.31 & 52.16 & 59.60 & 45.71 & 42.11 & 46.27 & 25.30 & 27.44 & 19.86 & 38.10 & 59.05 & 46.19 & 43.93 \\ \midrule
        &Avg. & 72.63 & 54.55 & 61.61 & 51.81 & 44.01 & 47.22 & 26.46 & 29.97 & 55.09 & 36.49 & 59.45 & 53.44 & 49.39 \\ 

\midrule 
\midrule

\multirow{15}{*}{\shortstack[l]{Supervised \\ Instruction-tuning}} & GPT-4 & \textbf{93.75} & \textbf{89.53} & \textbf{85.18} & \textbf{81.46} & \textbf{79.22} & \textbf{77.75} & \textbf{72.83} & \textbf{88.12} & \textbf{61.90} & \textbf{75.39} & \textbf{90.22} & \textbf{83.15} & \textbf{81.54} \\ 
         & ChatGPT & \underline{86.10} & \underline{78.58} & \underline{76.69} & \underline{79.15} & \underline{65.53} & 70.72 & 54.05 & \underline{79.82} & 53.12 & \underline{66.10} & 73.70 & 61.50 & \underline{70.42} \\ 
         & Baichuan2-13B-Chat & 77.65 & 73.09 & 67.15 & 76.71 & 61.50 & 66.69 & \underline{54.56} & 64.04 & \underline{55.80} & 56.52 & 72.24 & \underline{71.33} & 66.44 \\ 
         & InternLM-Chat-7B & 74.39 & 74.09 & 74.30 & 77.29 & 58.17 & \underline{71.19} & 45.80 & 67.54 & 51.84 & 60.40 & 72.11 & 45.62 & 64.40 \\ 
         & Qwen-7B-Chat & 74.62 & 73.09 & 65.43 & 76.57 & 62.05 & 65.17 & 48.20 & 66.79 & 49.66 & 49.50 & \underline{74.01} & 57.63 & 63.56 \\ 
         & ChatGLM2-6B & 68.92 & 65.34 & 67.55 & 67.29 & 60.80 & 63.56 & 45.44 & 53.76 & 48.29 & 39.97 & 66.26 & 56.64 & 58.65 \\ 
         & Baichaun-13B-Chat & 74.51 & 66.89 & 52.85 & 69.00 & 56.09 & 63.81 & 45.20 & 46.87 & 49.80 & 45.36 & 62.86 & 50.00 & 56.94 \\ 
         & Mistral-7B-Instruct & 57.97 & 59.68 & 70.19 & 69.00 & 54.47 & 62.71 & 41.72 & 30.07 & 40.62 & 45.98 & 72.78 & 54.27 & 54.96 \\
         & Vicuna-13B & 59.95 & 45.63 & 40.00 & 62.00 & 44.46 & 44.93 & 31.97 & 30.26 & 42.26 & 32.63 & 61.22 & 37.43 & 44.40 \\ 
         & Chinese Alpaca2-13B & 57.51 & 52.71 & 37.09 & 52.86 & 50.83 & 28.39 & 22.46 & 45.61 & 41.88 & 48.50 & 53.47 & 20.34 & 42.64 \\ 
         & LLaMA2-Chat-7B & 42.61 & 40.97 & 54.17 & 49.01 & 36.43 & 43.81 & 26.48 & 23.31 & 28.24 & 31.70 & 45.44 & 50.85 & 39.42 \\ 
         & MOSS-Moon-003-SFT & 32.48 & 35.44 & 45.30 & 47.14 & 28.53 & 41.61 & 21.17 & 3.26 & 13.51 & 33.21 & 48.57 & 32.77 & 31.92 \\ 
         & Xwin-7B & 35.04 & 29.90 & 30.60 & 37.14 & 26.32 & 29.49 & 25.30 & 14.29 & 24.42 & 25.94 & 41.36 & 15.54 & 27.95 \\ 
         \midrule
         & Avg. & 64.27 & 60.38 & 58.96 & 64.97 & 52.65 & 56.14 & 41.17 & 47.21 & 43.18 & 47.02 & 64.17 & 49.01 & 54.09 \\ \bottomrule

\end{tabular}}
\caption{Accuracy on Chinese DialogBench.
\textbf{Bold} and \underline{underlined} indicate the best results and second-best results.}
\label{tab:main_result}
\end{table*}
\paragraph{Evaluation Method.}
For the above LLMs, we use accuracy as the metric and adopt different evaluation methods.
(1) \textbf{Pre-trained LLMs:} 
each option content is independently scored by concatenating it with the instruction along with the given dialogue and question as a prompt 
and computing the probability of ``option content''. Specifically, we calculate the perplexity of each option content and then choose the label corresponding to the option content with the lowest perplexity as the predicted answer.
This evaluation method is consistent with the training method of pre-trained LLMs (i.e., next token prediction), stimulating the optimal performance of LLMs.
(2) \textbf{Supervised instruction-tuning LLMs:} 
We regard the given dialogue as the history of chatting between the user and the LLM. In the current interaction turn, we concatenate the instruction, along with the question and all options to form an exact string as the user's question to the LLM, and then the LLM gives the option label.
In implementation, we allow LLMs to output at most $256$ tokens, and then extract the outputted label as the predicted answer.

\paragraph{Implementation Details.}
We further describe the parameter settings for GPT-4 when generating data and LLMs to be evaluated.
When using GPT-4 to generate evaluation instances, we set temperature to 1, presence\_penalty to $0.6$, frequency\_penalty to $0$, and other parameters to default for the API parameters of GPT-4.
When evaluating LLMs, we set the temperature to 0, the presence\_penalty to 0.6, and the frequency\_penalty to 0 for ChatGPT and GPT-4.
Besides, the temperature is set to 0, max\_new\_tokens is set to 256, and other parameters to default for other open-source models.
Furthermore, the versions of ChatGPT and GPT-4 we use in our work are \texttt{gpt-3.5-turbo-0613} and \texttt{gpt-4-0314}. 
We implement our code using Pytorch\footnote{\url{https://pytorch.org/}} and Huggingface\footnote{\url{https://huggingface.co/}} and experiment on A100 80GB GPUs, spending an average of 20 minutes to 2 hours on each task while inferring with open-source models.
We also show the evaluation prompts in Appendix~\ref{appendix:prompt_setup}.

\section{Main Results}

Overall and task-specific scores on English and Chinese DialogBench are reported in Table~\ref{tab:main_result_en} and~\ref{tab:main_result}. 
The overall score of all LLMs on English DialogBench is slightly better than the score on Chinese DialogBench.
Additionally, the overall performance of all LLMs on each task generally has the same trend on English and Chinese DialogBench.

\paragraph{Pretrained LLMs.}
On this challenging benchmark, surprisingly we discover that some pre-trained LLMs (e.g., LLaMA2-70B in English DialogBench and Baichuan2-13B in Chinese DialogBench) have pretty good performances.
For other pre-trained LLMs, 
there is still much room for improvement in those fine-grained capabilities related to the human likeness.

\begin{table*}[t!]
\centering
\scriptsize
\begin{tabular}{@{}lccccccccccccc@{}}
    \toprule

    \multirow{2}{*}{\textbf{Method}} & \multicolumn{4}{c}{\textbf{Correctness}} & \multicolumn{2}{c}{\textbf{Coherence}} & \multicolumn{5}{c}{\textbf{Consistency}} & \multicolumn{1}{c}{\textbf{Safety}}\\ 
    \cmidrule(lr){2-5} \cmidrule(lr){6-7} \cmidrule(lr){8-12} \cmidrule(lr){13-13}

& \textbf{SF} & \textbf{IC} & \textbf{KRG} & \textbf{CRG} & \textbf{DI} & \textbf{MRG} & \textbf{PRG} & \textbf{RC} & \textbf{ED} & \textbf{NLI} & \textbf{DS} & \textbf{OD} & \textbf{Overall} \\  \midrule

    Optimized Prompt & 93.75 & 89.53 & 85.18 & 81.46 & 79.22 & 77.75 & 72.83 & 88.12 & 61.90 & 75.39 & 90.22 & 83.15 & 81.54 \\
    \midrule
    \, -Styles & 94.26 & 89.79 & 89.95 & 81.80 & 89.39 & 91.03 & 73.51 & 89.22 & 73.43 & 75.42 & 91.32 & 84.55 & 85.31 \\
\, -Filter & 87.29 & 89.22 & 81.81 & 81.12 & 73.05 & 74.97 & 72.39 & 80.27 & 55.14 & 71.97 & 89.12 & 70.88 & 77.27 \\

    \bottomrule
\end{tabular}
\caption{Ablation study on different components of our optimized prompt on GPT-4. 
}
\label{tab:ablation_result}
\end{table*}

We further observe that: 
(1) For correctness, most pre-trained LLMs can perform well on slot filling (SF) but are relatively poor on the other $3$ tasks.
(2) For personalization consistency, pre-trained LLMs as a whole have good performances in emotion perception (ED), whereas poor performance in personality following (PRG).
For semantic consistency, the decent performance on dialogue summarization (DS) indicates that pre-trained LLMs perform well in maintaining semantic alignment.
However, it is still relatively difficult in scenarios that require one-step reasoning, as shown by the performance on dialogue NLI.
(3) For coherence, the average performance of LLMs on dialogue infilling (DI) and multi-turn response generation (MRG) is relatively similar, and there is still much room for improvement.
(4) For offensive detection (OD), most pre-trained LLMs can empower a certain capability of offensive detection.
Overall, current pre-trained LLMs perform relatively well on correctness-related tasks and have greater room for improvement on tasks related to coherence and safety. For consistency-related tasks, pre-trained LLMs must be continuously optimized to possess corresponding high-order capabilities.

\begin{figure}
\raggedleft
\includegraphics[width=\linewidth,keepaspectratio]{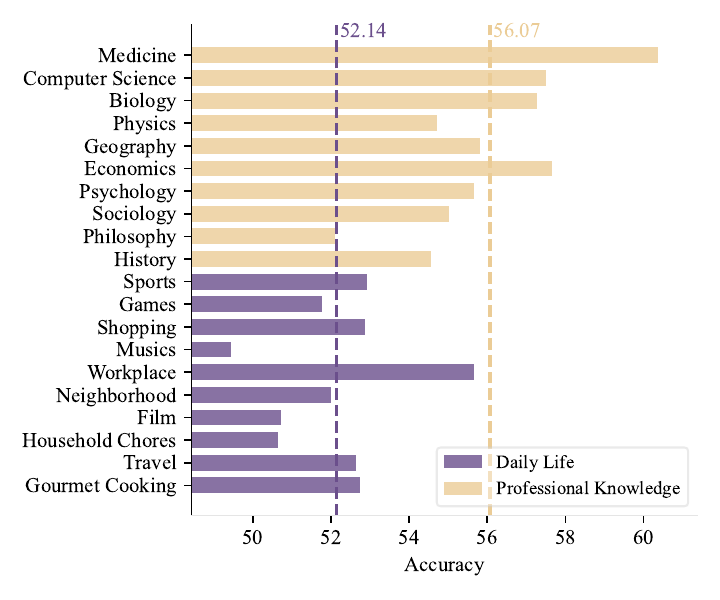}
\caption{The average accuracy on all supervised instruction-tuning LLMs for each domain. 
}

\label{fig:average_domain}
\end{figure}
\paragraph{Supervised Instruction-tuning LLMs.}
We begin by observing that GPT-4 presents the best performance, which represents the strongest capabilities as a human-like dialogue system.
Additionally, the instruction-tuning LLMs achieve higher scores than the corresponding pre-trained LLMs on most dialogue tasks (e.g., Baichuan2-13B), suggesting that instruction tuning is an efficient means for improving the capabilities that LLMs as human-like dialogue systems should have.

We further observe that:
(1) For correctness, most LLMs perform relatively well on all $4$ tasks, indicating that these LLMs have decent abilities to generate correct dialogues.
(2) For personalization consistency, most LLMs perform unsatisfactorily.
Interestingly, most LLMs achieve inferior scores on emotion classification than the corresponding pre-trained LLMs, such as QWen-7B.
It might be because the positioning of assistant AI enables instruction tuning to focus on the ability to complete tasks, abandoning the ability to perceive emotions.
(3) For coherence and safety, although instruction tuning has enhanced the LLMs' abilities, there is still much room for improvement.
Overall, there is the same trend on different evaluation dimensions for supervised instruction-tuning LLMs as pre-trained LLMs.
Due to space limitations, a more detailed experimental analysis is in Appendix~\ref{appendix:main_results}.

\section{Further Discussion}
We probe LLMs' performance for different domains and validate the effectiveness of 
adjusting dialogue style and introducing filtering mechanisms.

\paragraph{Performance on Different Domains.}
We calculate the average accuracy of all supervised instruction-tuning LLMs on each domain, as shown in Figure~\ref{fig:average_domain}.
The detailed results are displayed in Table~\ref{tab:domain_description}.
We observe that the average performance in daily life is overall lower than that in professional knowledge (e.g., $52.14\%$ vs. $56.07\%$).
We speculate that this is related to the current positioning of supervised instruction-tuning as assistant AI.
Assistant AI needs to follow instructions to complete various knowledge-based tasks, which particularly requires LLMs to master a variety of professional knowledge.
Correspondingly, information relevant to the daily life of humans might be underestimated when fine-tuning LLMs.
This suggests that improving the human-likeness of LLMs as dialogue systems requires introducing more daily dialogues into supervised fine-tuning.

\paragraph{Ablation Study.}
We perform the following ablation tests to validate the effect of each component:
(1) Remove the description of mitigating the style bias in the prompt (-Styles);
(2) Remove the filter mechanism (-Filter).
We use GPT-4 to conduct this experiment.
The results are shown in Table~\ref{tab:ablation_result}. 
We observe that: (1) The accuracy improves to varying degrees without mitigating the style bias, which validates that unfriendly communication would greatly increase the difficulty of interaction.
(2) The accuracy has dropped to varying degrees, indicating that the filtered instances are indeed incorrect and LLMs cannot answer.

\section{Conclusion}
We present DialogBench, a systematically designed dialogue benchmark for evaluating LLMs as human-like dialogue systems.
DialogBench includes $12$ dialogue tasks to probe the capabilities related to human likeness for comprehensive evaluation.
For each task, we prompt GPT-4 to generate evaluation instances.
Specifically, we design the basic prompt based on widely-used design principles and further eliminate existing biases to generate higher-quality instances.
An extensive study of $26$ LLMs, including pre-trained and supervised instruction-tuning, is conducted in Chinese and English DialogBench.
We unveil that instruction fine-tuning can improve the human likeness of LLMs to a certain extent.
However, there is still a long way to go for most LLMs as human-like dialogue systems.
In addition, LLMs are generally better at understanding context, but relatively poor at perceiving emotions and personality.
We expect DialogBench to serve as a cornerstone for future study to develop better human-like LLMs.

\section*{Limitations}

\paragraph{Multilingual Benchmark Expansions.}
DialogBench can only be used to evaluate English and Chinese LLMs, and cannot evaluate LLMs in other languages. 
However, our proposed evaluation framework is applicable to all LLM evaluations, which only need to use the top-tier LLM of the corresponding language as a data generator to construct evaluation instances for quickly building a testbed.

\paragraph{Additional Dimensions and Dialogue Tasks.}
Human-like dialogue systems require a variety of fine-grained capabilities to ensure long-term connections with users. Although we conduct extensive literature references to select comprehensive dimensions and dialogue tasks, we fully acknowledge that some other dimensions and dialogue tasks were not included in our benchmark.
In addition, we employ GPT-4 as a data generator, which sets restrictions on the selection of dimensions and dialogue tasks.
some researchers~\citep{chang2023survey,bubeck2023sparks} have highlighted clear limitations of GPT-4, including limited reasoning, output length limit, and toxic content generation.
Therefore, we pay more attention to dimensions and dialogue tasks that GPT-4 experts in.

\paragraph{Technical limitations.}
Due to limited computational and financial resources, we only include pre-trained LLMs with no more than $70$B and supervised instruction-tuning LLMs with no more than $20$B in DialogBench’s first edition of evaluation.
Although recent research suggests that when LLMs expand beyond a certain threshold, they may begin to exhibit emerging capabilities~\cite{wei2022emergent}, we were unable to test all very large language models.
We welcome future researchers to study our benchmarks and evaluate LLMs as human-like dialogue systems.

\paragraph{Reproducibility of Closed Access Models.}
Some of the LLMs (e.g., ChatGPT and GPT-4) being evaluated are only accessible through a programming interface that essentially adds a black box on top of a black box. 
The mechanisms behind these interfaces may change at any time, so the evaluation results from different periods may vary arbitrarily.

\section*{Ethics Statement}

Since GPT-4 is trained on online data, GPT-4 may encode biases that perpetuate stereotypes, discrimination, or marginalization of specific languages or communities. 
This results in DialogBench potentially generating toxic and harmful instances
Furthermore, we induce GPT-4 to generate a certain proportion of unfriendly dialogues for evaluating LLMs in unfriendly scenarios, which can reflect the true level of LLMs as human-like dialogue systems.
Accordingly, this might lead to some unkind and harmful instances.
In addition, we employ three experts to manually do these evaluation questions. We pay $0.2$ to each expert for each instance.

\section*{Acknowledgements}
We sincerely thank the anonymous reviewers for their thorough reviewing and valuable suggestions.

\bibliography{custom}  

\begin{thebibliography}{70}
\expandafter\ifx\csname natexlab\endcsname\relax\def\natexlab#1{#1}\fi

\bibitem[{Acheampong et~al.(2020)Acheampong, Wenyu, and Nunoo-Mensah}]{acheampong2020text}
Francisca~Adoma Acheampong, Chen Wenyu, and Henry Nunoo-Mensah. 2020.
\newblock Text-based emotion detection: Advances, challenges, and opportunities.
\newblock \emph{Engineering Reports}, 2(7):e12189.

\bibitem[{Ahn et~al.(2022)Ahn, Brohan, Brown, Chebotar, Cortes, David, Finn, Fu, Gopalakrishnan, Hausman et~al.}]{ahn2022can}
Michael Ahn, Anthony Brohan, Noah Brown, Yevgen Chebotar, Omar Cortes, Byron David, Chelsea Finn, Chuyuan Fu, Keerthana Gopalakrishnan, Karol Hausman, et~al. 2022.
\newblock Do as i can, not as i say: Grounding language in robotic affordances.
\newblock \emph{arXiv preprint arXiv:2204.01691}.

\bibitem[{Almazrouei et~al.(2023)Almazrouei, Alobeidli, Alshamsi, Cappelli, Cojocaru, Debbah, Goffinet, Heslow, Launay, Malartic, Noune, Pannier, and Penedo}]{falcon40b}
Ebtesam Almazrouei, Hamza Alobeidli, Abdulaziz Alshamsi, Alessandro Cappelli, Ruxandra Cojocaru, Merouane Debbah, Etienne Goffinet, Daniel Heslow, Julien Launay, Quentin Malartic, Badreddine Noune, Baptiste Pannier, and Guilherme Penedo. 2023.
\newblock {Falcon-40B}: an open large language model with state-of-the-art performance.

\bibitem[{Babe et~al.(2023)Babe, Nguyen, Zi, Guha, Feldman, and Anderson}]{babe2023studenteval}
Hannah~McLean Babe, Sydney Nguyen, Yangtian Zi, Arjun Guha, Molly~Q Feldman, and Carolyn~Jane Anderson. 2023.
\newblock Studenteval: A benchmark of student-written prompts for large language models of code.
\newblock \emph{arXiv preprint arXiv:2306.04556}.

\bibitem[{Bai et~al.(2023)Bai, Bai, Chu, Cui, Dang, Deng, Fan, Ge, Han, Huang, Hui, Ji, Li, Lin, Lin, Liu, Liu, Lu, Lu, Ma, Men, Ren, Ren, Tan, Tan, Tu, Wang, Wang, Wang, Wu, Xu, Xu, Yang, Yang, Yang, Yang, Yang, Yao, Yu, Bowen, Yuan, Yuan, Zhang, Zhang, Zhang, Zhang, Zhou, Zhou, Zhou, and Zhu}]{Bai2023QwenTR}
Jinze Bai, Shuai Bai, Yunfei Chu, Zeyu Cui, Kai Dang, Xiaodong Deng, Yang Fan, Wenhang Ge, Yu~Han, Fei Huang, Binyuan Hui, Luo Ji, Mei Li, Junyang Lin, Runji Lin, Dayiheng Liu, Gao Liu, Chengqiang Lu, K.~Lu, Jianxin Ma, Rui Men, Xingzhang Ren, Xuancheng Ren, Chuanqi Tan, Sinan Tan, Jianhong Tu, Peng Wang, Shijie Wang, Wei Wang, Shengguang Wu, Benfeng Xu, Jin Xu, An~Yang, Hao Yang, Jian Yang, Jian Yang, Shusheng Yang, Yang Yao, Bowen Yu, Yu~Bowen, Hongyi Yuan, Zheng Yuan, Jianwei Zhang, Xing Zhang, Yichang Zhang, Zhenru Zhang, Chang Zhou, Jingren Zhou, Xiaohuan Zhou, and Tianhang Zhu. 2023.
\newblock \href {https://api.semanticscholar.org/CorpusID:263134555} {Qwen technical report}.
\newblock \emph{ArXiv}, abs/2309.16609.

\bibitem[{Bai et~al.(2022)Bai, Jones, Ndousse, Askell, Chen, DasSarma, Drain, Fort, Ganguli, Henighan et~al.}]{bai2022training}
Yuntao Bai, Andy Jones, Kamal Ndousse, Amanda Askell, Anna Chen, Nova DasSarma, Dawn Drain, Stanislav Fort, Deep Ganguli, Tom Henighan, et~al. 2022.
\newblock Training a helpful and harmless assistant with reinforcement learning from human feedback.
\newblock \emph{arXiv preprint arXiv:2204.05862}.

\bibitem[{Bubeck et~al.(2023)Bubeck, Chandrasekaran, Eldan, Gehrke, Horvitz, Kamar, Lee, Lee, Li, Lundberg et~al.}]{bubeck2023sparks}
S{\'e}bastien Bubeck, Varun Chandrasekaran, Ronen Eldan, Johannes Gehrke, Eric Horvitz, Ece Kamar, Peter Lee, Yin~Tat Lee, Yuanzhi Li, Scott Lundberg, et~al. 2023.
\newblock Sparks of artificial general intelligence: Early experiments with gpt-4.
\newblock \emph{arXiv preprint arXiv:2303.12712}.

\bibitem[{Chalamalasetti et~al.(2023)Chalamalasetti, G{\"o}tze, Hakimov, Madureira, Sadler, and Schlangen}]{chalamalasetti-etal-2023-clembench}
Kranti Chalamalasetti, Jana G{\"o}tze, Sherzod Hakimov, Brielen Madureira, Philipp Sadler, and David Schlangen. 2023.
\newblock clembench: Using game play to evaluate chat-optimized language models as conversational agents.
\newblock In \emph{Proceedings of the 2023 Conference on Empirical Methods in Natural Language Processing}, pages 11174--11219, Singapore.

\bibitem[{Chang et~al.(2023)Chang, Wang, Wang, Wu, Zhu, Chen, Yang, Yi, Wang, Wang et~al.}]{chang2023survey}
Yupeng Chang, Xu~Wang, Jindong Wang, Yuan Wu, Kaijie Zhu, Hao Chen, Linyi Yang, Xiaoyuan Yi, Cunxiang Wang, Yidong Wang, et~al. 2023.
\newblock A survey on evaluation of large language models.
\newblock \emph{arXiv preprint arXiv:2307.03109}.

\bibitem[{Chen et~al.(2017)Chen, Liu, Yin, and Tang}]{chen2017survey}
Hongshen Chen, Xiaorui Liu, Dawei Yin, and Jiliang Tang. 2017.
\newblock A survey on dialogue systems: Recent advances and new frontiers.
\newblock \emph{Acm Sigkdd Explorations Newsletter}, 19(2):25--35.

\bibitem[{Cheng et~al.(2023)Cheng, Li, and Bing}]{cheng2023gpt}
Liying Cheng, Xingxuan Li, and Lidong Bing. 2023.
\newblock Is gpt-4 a good data analyst?
\newblock \emph{arXiv preprint arXiv:2305.15038}.

\bibitem[{Choi et~al.(2023)Choi, Pei, Kumar, Shu, and Jurgens}]{choi-etal-2023-llms}
Minje Choi, Jiaxin Pei, Sagar Kumar, Chang Shu, and David Jurgens. 2023.
\newblock Do {LLM}s understand social knowledge? evaluating the sociability of large language models with {S}oc{KET} benchmark.
\newblock In \emph{Proceedings of the 2023 Conference on Empirical Methods in Natural Language Processing}, pages 11370--11403, Singapore.

\bibitem[{Clark et~al.()Clark, Cowhey, Etzioni, Khot, Sabharwal, Schoenick, and Tafjord}]{clarkthink}
Peter Clark, Isaac Cowhey, Oren Etzioni, Tushar Khot, Ashish Sabharwal, Carissa Schoenick, and Oyvind Tafjord.
\newblock Think you have solved question answering? try arc, the ai2 reasoning challenge.

\bibitem[{Cobbe et~al.(2021)Cobbe, Kosaraju, Bavarian, Chen, Jun, Kaiser, Plappert, Tworek, Hilton, Nakano et~al.}]{cobbe2021training}
Karl Cobbe, Vineet Kosaraju, Mohammad Bavarian, Mark Chen, Heewoo Jun, Lukasz Kaiser, Matthias Plappert, Jerry Tworek, Jacob Hilton, Reiichiro Nakano, et~al. 2021.
\newblock Training verifiers to solve math word problems.
\newblock \emph{arXiv preprint arXiv:2110.14168}.

\bibitem[{Cui et~al.(2023)Cui, Yang, and Yao}]{Cui2023EfficientAE}
Yiming Cui, Ziqing Yang, and Xin Yao. 2023.
\newblock \href {https://api.semanticscholar.org/CorpusID:258180548} {Efficient and effective text encoding for chinese llama and alpaca}.
\newblock \emph{ArXiv}, abs/2304.08177.

\bibitem[{Dinan et~al.(2019)Dinan, Humeau, Chintagunta, and Weston}]{dinan2019build}
Emily Dinan, Samuel Humeau, Bharath Chintagunta, and Jason Weston. 2019.
\newblock Build it break it fix it for dialogue safety: Robustness from adversarial human attack.
\newblock In \emph{Proceedings of the 2019 Conference on Empirical Methods in Natural Language Processing and the 9th International Joint Conference on Natural Language Processing (EMNLP-IJCNLP)}, pages 4537--4546.

\bibitem[{Du et~al.(2022)Du, Qian, Liu, Ding, Qiu, Yang, and Tang}]{du2022glm}
Zhengxiao Du, Yujie Qian, Xiao Liu, Ming Ding, Jiezhong Qiu, Zhilin Yang, and Jie Tang. 2022.
\newblock Glm: General language model pretraining with autoregressive blank infilling.
\newblock In \emph{Proceedings of the 60th Annual Meeting of the Association for Computational Linguistics (Volume 1: Long Papers)}, pages 320--335.

\bibitem[{Feng et~al.()Feng, Feng, and Qin}]{fengsurvey}
Xiachong Feng, Xiaocheng Feng, and Bing Qin.
\newblock A survey on dialogue summarization: Recent advances and new frontiers.

\bibitem[{Gupta et~al.(2022)Gupta, Jiao, Yeh, Mehri, Eskenazi, and Bigham}]{gupta-etal-2022-instructdial}
Prakhar Gupta, Cathy Jiao, Yi-Ting Yeh, Shikib Mehri, Maxine Eskenazi, and Jeffrey Bigham. 2022.
\newblock \href {https://doi.org/10.18653/v1/2022.emnlp-main.33} {{I}nstruct{D}ial: Improving zero and few-shot generalization in dialogue through instruction tuning}.
\newblock In \emph{Proceedings of the 2022 Conference on Empirical Methods in Natural Language Processing}, pages 505--525, Abu Dhabi, United Arab Emirates. Association for Computational Linguistics.

\bibitem[{Hendrycks et~al.(2021)Hendrycks, Burns, Basart, Zou, Mazeika, Song, and Steinhardt}]{hendrycks2021measuring}
Dan Hendrycks, Collin Burns, Steven Basart, Andy Zou, Mantas Mazeika, Dawn Song, and Jacob Steinhardt. 2021.
\newblock \href {https://openreview.net/forum?id=d7KBjmI3GmQ} {Measuring massive multitask language understanding}.
\newblock In \emph{International Conference on Learning Representations}.

\bibitem[{Huang et~al.(2020)Huang, Zhu, and Gao}]{huang2020challenges}
Minlie Huang, Xiaoyan Zhu, and Jianfeng Gao. 2020.
\newblock Challenges in building intelligent open-domain dialog systems.
\newblock \emph{ACM Transactions on Information Systems (TOIS)}, 38(3):1--32.

\bibitem[{Huang et~al.(2023)Huang, Bai, Zhu, Zhang, Zhang, Su, Liu, Lv, Zhang, Lei et~al.}]{huang2023c}
Yuzhen Huang, Yuzhuo Bai, Zhihao Zhu, Junlei Zhang, Jinghan Zhang, Tangjun Su, Junteng Liu, Chuancheng Lv, Yikai Zhang, Jiayi Lei, et~al. 2023.
\newblock C-eval: A multi-level multi-discipline chinese evaluation suite for foundation models.
\newblock \emph{arXiv preprint arXiv:2305.08322}.

\bibitem[{Ji et~al.(2023)Ji, Wu, Zheng, Hu, Chen, and He}]{ji2023chatgpt}
Yu~Ji, Wen Wu, Hong Zheng, Yi~Hu, Xi~Chen, and Liang He. 2023.
\newblock Is chatgpt a good personality recognizer? a preliminary study.
\newblock \emph{arXiv preprint arXiv:2307.03952}.

\bibitem[{Jia et~al.(2021)Jia, Huang, and Zhu}]{jia2021ddrel}
Qi~Jia, Hongru Huang, and Kenny~Q Zhu. 2021.
\newblock Ddrel: A new dataset for interpersonal relation classification in dyadic dialogues.
\newblock In \emph{Proceedings of the AAAI Conference on Artificial Intelligence}, volume~35, pages 13125--13133.

\bibitem[{Jiang et~al.(2023)Jiang, Sablayrolles, Mensch, Bamford, Chaplot, Casas, Bressand, Lengyel, Lample, Saulnier et~al.}]{jiang2023mistral}
Albert~Q Jiang, Alexandre Sablayrolles, Arthur Mensch, Chris Bamford, Devendra~Singh Chaplot, Diego de~las Casas, Florian Bressand, Gianna Lengyel, Guillaume Lample, Lucile Saulnier, et~al. 2023.
\newblock Mistral 7b.
\newblock \emph{arXiv preprint arXiv:2310.06825}.

\bibitem[{Li et~al.(2023{\natexlab{a}})Li, Zhang, Koto, Yang, Zhao, Gong, Duan, and Baldwin}]{li2023cmmlu}
Haonan Li, Yixuan Zhang, Fajri Koto, Yifei Yang, Hai Zhao, Yeyun Gong, Nan Duan, and Timothy Baldwin. 2023{\natexlab{a}}.
\newblock \href {http://arxiv.org/abs/2306.09212} {Cmmlu: Measuring massive multitask language understanding in chinese}.

\bibitem[{Li et~al.(2023{\natexlab{b}})Li, Zhao, Yu, Song, Li, Yu, Li, Huang, and Li}]{li-etal-2023-api}
Minghao Li, Yingxiu Zhao, Bowen Yu, Feifan Song, Hangyu Li, Haiyang Yu, Zhoujun Li, Fei Huang, and Yongbin Li. 2023{\natexlab{b}}.
\newblock {API}-bank: A comprehensive benchmark for tool-augmented {LLM}s.
\newblock pages 3102--3116, Singapore. Association for Computational Linguistics.

\bibitem[{Li et~al.(2017)Li, Su, Shen, Li, Cao, and Niu}]{li2017dailydialog}
Yanran Li, Hui Su, Xiaoyu Shen, Wenjie Li, Ziqiang Cao, and Shuzi Niu. 2017.
\newblock Dailydialog: A manually labelled multi-turn dialogue dataset.
\newblock In \emph{Proceedings of the Eighth International Joint Conference on Natural Language Processing (Volume 1: Long Papers)}, pages 986--995.

\bibitem[{Liang et~al.(2022)Liang, Bommasani, Lee, Tsipras, Soylu, Yasunaga, Zhang, Narayanan, Wu, Kumar et~al.}]{liang2022holistic}
Percy Liang, Rishi Bommasani, Tony Lee, Dimitris Tsipras, Dilara Soylu, Michihiro Yasunaga, Yian Zhang, Deepak Narayanan, Yuhuai Wu, Ananya Kumar, et~al. 2022.
\newblock Holistic evaluation of language models.
\newblock \emph{arXiv preprint arXiv:2211.09110}.

\bibitem[{Liu et~al.(2023)Liu, Yu, Zhang, Xu, Lei, Lai, Gu, Ding, Men, Yang et~al.}]{liu2023agentbench}
Xiao Liu, Hao Yu, Hanchen Zhang, Yifan Xu, Xuanyu Lei, Hanyu Lai, Yu~Gu, Hangliang Ding, Kaiwen Men, Kejuan Yang, et~al. 2023.
\newblock Agentbench: Evaluating llms as agents.
\newblock \emph{arXiv preprint arXiv:2308.03688}.

\bibitem[{Louvan and Magnini(2020)}]{louvan2020recent}
Samuel Louvan and Bernardo Magnini. 2020.
\newblock Recent neural methods on slot filling and intent classification for task-oriented dialogue systems: A survey.
\newblock In \emph{Proceedings of the 28th International Conference on Computational Linguistics}, pages 480--496.

\bibitem[{Ma et~al.(2020)Ma, Nguyen, Xing, and Cambria}]{ma2020survey}
Yukun Ma, Khanh~Linh Nguyen, Frank~Z Xing, and Erik Cambria. 2020.
\newblock A survey on empathetic dialogue systems.
\newblock \emph{Information Fusion}, 64:50--70.

\bibitem[{Mehri et~al.(2020)Mehri, Eric, and Hakkani-Tur}]{mehri2020dialoglue}
Shikib Mehri, Mihail Eric, and Dilek Hakkani-Tur. 2020.
\newblock Dialoglue: A natural language understanding benchmark for task-oriented dialogue.
\newblock \emph{arXiv preprint arXiv:2009.13570}.

\bibitem[{Mehri and Eskenazi(2020)}]{mehri-eskenazi-2020-usr}
Shikib Mehri and Maxine Eskenazi. 2020.
\newblock {USR}: An unsupervised and reference free evaluation metric for dialog generation.
\newblock In \emph{Proceedings of the 58th Annual Meeting of the Association for Computational Linguistics}, pages 681--707, Online. Association for Computational Linguistics.

\bibitem[{Mishra et~al.(2022)Mishra, Khashabi, Baral, and Hajishirzi}]{mishra2022cross}
Swaroop Mishra, Daniel Khashabi, Chitta Baral, and Hannaneh Hajishirzi. 2022.
\newblock Cross-task generalization via natural language crowdsourcing instructions.
\newblock In \emph{Proceedings of the 60th Annual Meeting of the Association for Computational Linguistics (Volume 1: Long Papers)}, pages 3470--3487.

\bibitem[{M{\o}ller et~al.(2023)M{\o}ller, Dalsgaard, Pera, and Aiello}]{moller2023prompt}
Anders~Giovanni M{\o}ller, Jacob~Aarup Dalsgaard, Arianna Pera, and Luca~Maria Aiello. 2023.
\newblock Is a prompt and a few samples all you need? using gpt-4 for data augmentation in low-resource classification tasks.
\newblock \emph{arXiv preprint arXiv:2304.13861}.

\bibitem[{OpenAI(2022)}]{chatgpt}
OpenAI. 2022.
\newblock \href {https://openai.com/blog/chatgpt/} {Chatgpt: Optimizing language models for dialogue}.
\newblock \emph{OpenAI Blog}.

\bibitem[{OpenAI(2023)}]{OpenAI2023GPT4TR}
OpenAI. 2023.
\newblock Gpt-4 technical report.
\newblock \emph{ArXiv}, abs/2303.08774.

\bibitem[{Rao et~al.(2023)Rao, Leung, and Miao}]{rao2023can}
Haocong Rao, Cyril Leung, and Chunyan Miao. 2023.
\newblock Can chatgpt assess human personalities? a general evaluation framework.
\newblock \emph{arXiv preprint arXiv:2303.01248}.

\bibitem[{Reddy et~al.(2019)Reddy, Chen, and Manning}]{reddy-etal-2019-coqa}
Siva Reddy, Danqi Chen, and Christopher~D. Manning. 2019.
\newblock \href {https://doi.org/10.1162/tacl_a_00266} {{C}o{QA}: A conversational question answering challenge}.
\newblock \emph{Transactions of the Association for Computational Linguistics}, 7:249--266.

\bibitem[{Safdari et~al.(2023)Safdari, Serapio-Garc{\'\i}a, Crepy, Fitz, Romero, Sun, Abdulhai, Faust, and Matari{\'c}}]{safdari2023personality}
Mustafa Safdari, Greg Serapio-Garc{\'\i}a, Cl{\'e}ment Crepy, Stephen Fitz, Peter Romero, Luning Sun, Marwa Abdulhai, Aleksandra Faust, and Maja Matari{\'c}. 2023.
\newblock Personality traits in large language models.
\newblock \emph{arXiv preprint arXiv:2307.00184}.

\bibitem[{Santhanam et~al.(2021)Santhanam, Hedayatnia, Gella, Padmakumar, Kim, Liu, and Hakkani-T{\"u}r}]{santhanam2021rome}
Sashank Santhanam, Behnam Hedayatnia, Spandana Gella, Aishwarya Padmakumar, Seokhwan Kim, Yang Liu, and Dilek Hakkani-T{\"u}r. 2021.
\newblock Rome was built in 1776: A case study on factual correctness in knowledge-grounded response generation.

\bibitem[{Srivastava et~al.(2023)Srivastava, Rastogi, Rao, Shoeb, Abid, Fisch, Brown, Santoro, Gupta, Garriga-Alonso et~al.}]{srivastava2023beyond}
Aarohi Srivastava, Abhinav Rastogi, Abhishek Rao, Abu Awal~Md Shoeb, Abubakar Abid, Adam Fisch, Adam~R Brown, Adam Santoro, Aditya Gupta, Adri{\`a} Garriga-Alonso, et~al. 2023.
\newblock Beyond the imitation game: Quantifying and extrapolating the capabilities of language models.
\newblock \emph{Transactions on Machine Learning Research}.

\bibitem[{Sun et~al.(2023)Sun, Zhang, He, Li, Cheng, Yan, Liu, Shao, Tang, Zhao, Chen, Zheng, Zhou, Li, Zhan, Zhou, Li, Yang, Wu, Yin, Huang, and Qiu}]{sun2023moss}
Tianxiang Sun, Xiaotian Zhang, Zhengfu He, Peng Li, Qinyuan Cheng, Hang Yan, Xiangyang Liu, Yunfan Shao, Qiong Tang, Xingjian Zhao, Ke~Chen, Yining Zheng, Zhejian Zhou, Ruixiao Li, Jun Zhan, Yunhua Zhou, Linyang Li, Xiaogui Yang, Lingling Wu, Zhangyue Yin, Xuanjing Huang, and Xipeng Qiu. 2023.
\newblock Moss: Training conversational language models from synthetic data.

\bibitem[{Tang et~al.(2023)Tang, Han, Jiang, and Hu}]{tang2023does}
Ruixiang Tang, Xiaotian Han, Xiaoqian Jiang, and Xia Hu. 2023.
\newblock Does synthetic data generation of llms help clinical text mining?
\newblock \emph{arXiv preprint arXiv:2303.04360}.

\bibitem[{Team(2023{\natexlab{a}})}]{2023internlm}
InternLM Team. 2023{\natexlab{a}}.
\newblock Internlm: A multilingual language model with progressively enhanced capabilities.
\newblock \url{https://github.com/InternLM/InternLM}.

\bibitem[{Team(2023{\natexlab{b}})}]{xwin-lm}
Xwin-LM Team. 2023{\natexlab{b}}.
\newblock \href {https://github.com/Xwin-LM/Xwin-LM} {Xwin-lm}.

\bibitem[{Touvron et~al.(2023{\natexlab{a}})Touvron, Lavril, Izacard, Martinet, Lachaux, Lacroix, Rozi{\`e}re, Goyal, Hambro, Azhar, Rodriguez, Joulin, Grave, and Lample}]{Touvron2023LLaMAOA}
Hugo Touvron, Thibaut Lavril, Gautier Izacard, Xavier Martinet, Marie-Anne Lachaux, Timoth{\'e}e Lacroix, Baptiste Rozi{\`e}re, Naman Goyal, Eric Hambro, Faisal Azhar, Aurelien Rodriguez, Armand Joulin, Edouard Grave, and Guillaume Lample. 2023{\natexlab{a}}.
\newblock \href {https://api.semanticscholar.org/CorpusID:257219404} {Llama: Open and efficient foundation language models}.
\newblock \emph{ArXiv}, abs/2302.13971.

\bibitem[{Touvron et~al.(2023{\natexlab{b}})Touvron, Martin, Stone, Albert, Almahairi, Babaei, Bashlykov, Batra, Bhargava, Bhosale, Bikel, Blecher, Ferrer, Chen, Cucurull, Esiobu, Fernandes, Fu, Fu, Fuller, Gao, Goswami, Goyal, Hartshorn, Hosseini, Hou, Inan, Kardas, Kerkez, Khabsa, Kloumann, Korenev, Koura, Lachaux, Lavril, Lee, Liskovich, Lu, Mao, Martinet, Mihaylov, Mishra, Molybog, Nie, Poulton, Reizenstein, Rungta, Saladi, Schelten, Silva, Smith, Subramanian, Tan, Tang, Taylor, Williams, Kuan, Xu, Yan, Zarov, Zhang, Fan, Kambadur, Narang, Rodriguez, Stojnic, Edunov, and Scialom}]{Touvron2023Llama2O}
Hugo Touvron, Louis Martin, Kevin~R. Stone, Peter Albert, Amjad Almahairi, Yasmine Babaei, Nikolay Bashlykov, Soumya Batra, Prajjwal Bhargava, Shruti Bhosale, Daniel~M. Bikel, Lukas Blecher, Cristian~Cant{\'o}n Ferrer, Moya Chen, Guillem Cucurull, David Esiobu, Jude Fernandes, Jeremy Fu, Wenyin Fu, Brian Fuller, Cynthia Gao, Vedanuj Goswami, Naman Goyal, Anthony~S. Hartshorn, Saghar Hosseini, Rui Hou, Hakan Inan, Marcin Kardas, Viktor Kerkez, Madian Khabsa, Isabel~M. Kloumann, A.~V. Korenev, Punit~Singh Koura, Marie-Anne Lachaux, Thibaut Lavril, Jenya Lee, Diana Liskovich, Yinghai Lu, Yuning Mao, Xavier Martinet, Todor Mihaylov, Pushkar Mishra, Igor Molybog, Yixin Nie, Andrew Poulton, Jeremy Reizenstein, Rashi Rungta, Kalyan Saladi, Alan Schelten, Ruan Silva, Eric~Michael Smith, R.~Subramanian, Xia Tan, Binh Tang, Ross Taylor, Adina Williams, Jian~Xiang Kuan, Puxin Xu, Zhengxu Yan, Iliyan Zarov, Yuchen Zhang, Angela Fan, Melanie Kambadur, Sharan Narang, Aurelien Rodriguez, Robert Stojnic, Sergey Edunov, and
  Thomas Scialom. 2023{\natexlab{b}}.
\newblock \href {https://api.semanticscholar.org/CorpusID:259950998} {Llama 2: Open foundation and fine-tuned chat models}.
\newblock \emph{ArXiv}, abs/2307.09288.

\bibitem[{Wang et~al.(2023{\natexlab{a}})Wang, Wang, Mi, Deng, Wang, Liang, Xu, and Wong}]{wang-etal-2023-cue}
Hongru Wang, Rui Wang, Fei Mi, Yang Deng, Zezhong Wang, Bin Liang, Ruifeng Xu, and Kam-Fai Wong. 2023{\natexlab{a}}.
\newblock Cue-{C}o{T}: Chain-of-thought prompting for responding to in-depth dialogue questions with {LLM}s.
\newblock In \emph{Findings of the Association for Computational Linguistics: EMNLP 2023}, pages 12047--12064, Singapore.

\bibitem[{Wang et~al.(2023{\natexlab{b}})Wang, Xie, Ding, Feng, and Xia}]{wang2023chatgpt}
Zengzhi Wang, Qiming Xie, Zixiang Ding, Yi~Feng, and Rui Xia. 2023{\natexlab{b}}.
\newblock Is chatgpt a good sentiment analyzer? a preliminary study.
\newblock \emph{arXiv preprint arXiv:2304.04339}.

\bibitem[{Wei et~al.(2021)Wei, Bosma, Zhao, Guu, Yu, Lester, Du, Dai, and Le}]{wei2021finetuned}
Jason Wei, Maarten Bosma, Vincent~Y Zhao, Kelvin Guu, Adams~Wei Yu, Brian Lester, Nan Du, Andrew~M Dai, and Quoc~V Le. 2021.
\newblock Finetuned language models are zero-shot learners.
\newblock \emph{arXiv preprint arXiv:2109.01652}.

\bibitem[{Wei et~al.(2022{\natexlab{a}})Wei, Tay, Bommasani, Raffel, Zoph, Borgeaud, Yogatama, Bosma, Zhou, Metzler et~al.}]{wei2022emergent}
Jason Wei, Yi~Tay, Rishi Bommasani, Colin Raffel, Barret Zoph, Sebastian Borgeaud, Dani Yogatama, Maarten Bosma, Denny Zhou, Donald Metzler, et~al. 2022{\natexlab{a}}.
\newblock Emergent abilities of large language models.
\newblock \emph{arXiv preprint arXiv:2206.07682}.

\bibitem[{Wei et~al.(2022{\natexlab{b}})Wei, Wang, Schuurmans, Bosma, Xia, Chi, Le, Zhou et~al.}]{wei2022chain}
Jason Wei, Xuezhi Wang, Dale Schuurmans, Maarten Bosma, Fei Xia, Ed~Chi, Quoc~V Le, Denny Zhou, et~al. 2022{\natexlab{b}}.
\newblock Chain-of-thought prompting elicits reasoning in large language models.
\newblock \emph{Advances in Neural Information Processing Systems}, 35:24824--24837.

\bibitem[{Welleck et~al.(2019)Welleck, Weston, Szlam, and Cho}]{welleck2019dialogue}
Sean Welleck, Jason Weston, Arthur Szlam, and Kyunghyun Cho. 2019.
\newblock Dialogue natural language inference.
\newblock In \emph{Proceedings of the 57th Annual Meeting of the Association for Computational Linguistics}, pages 3731--3741.

\bibitem[{Whitehouse et~al.(2023)Whitehouse, Choudhury, and Aji}]{whitehouse-etal-2023-llm}
Chenxi Whitehouse, Monojit Choudhury, and Alham Aji. 2023.
\newblock {LLM}-powered data augmentation for enhanced cross-lingual performance.
\newblock In \emph{Proceedings of the 2023 Conference on Empirical Methods in Natural Language Processing}, pages 671--686, Singapore.

\bibitem[{Xu et~al.(2023{\natexlab{a}})Xu, Guo, Duan, and McAuley}]{Xu2023BaizeAO}
Canwen Xu, Daya Guo, Nan Duan, and Julian McAuley. 2023{\natexlab{a}}.
\newblock \href {https://api.semanticscholar.org/CorpusID:257912848} {Baize: An open-source chat model with parameter-efficient tuning on self-chat data}.
\newblock \emph{ArXiv}, abs/2304.01196.

\bibitem[{Xu et~al.(2023{\natexlab{b}})Xu, Li, Zhu, Xue, Zhu, Zhao, He, Zhang, Kang, and Lan}]{xu2023superclue}
Liang Xu, Anqi Li, Lei Zhu, Hang Xue, Changtai Zhu, Kangkang Zhao, Haonan He, Xuanwei Zhang, Qiyue Kang, and Zhenzhong Lan. 2023{\natexlab{b}}.
\newblock Superclue: A comprehensive chinese large language model benchmark.
\newblock \emph{arXiv preprint arXiv:2307.15020}.

\bibitem[{Xue et~al.(2022)Xue, Takiguchi, and Ariki}]{xue2022building}
Qiang Xue, Tetsuya Takiguchi, and Yasuo Ariki. 2022.
\newblock Building a knowledge-based dialogue system with text infilling.
\newblock In \emph{Proceedings of the 23rd Annual Meeting of the Special Interest Group on Discourse and Dialogue}, pages 237--243.

\bibitem[{Yang et~al.(2023)Yang, Xiao, Wang, Zhang, Bian, Yin, Lv, Pan, Wang, Yan, Yang, Deng, Wang, Liu, Ai, Dong, Zhao, Xu, Sun, Zhang, Liu, Ji, Xie, Dai, Fang, Su, Song, Liu, Ru, Ma, Wang, Liu, Lin, Nie, Guo, Sun, Tao, Li, Li, Cheng, Chen, Zeng, Wang, Chen, Men, Yu, Pan, Shen, Wang, Li, Jiang, Gao, Zhang, Zhou, and Wu}]{Yang2023Baichuan2O}
Ai~Ming Yang, Bin Xiao, Bingning Wang, Borong Zhang, Ce~Bian, Chao Yin, Chenxu Lv, Da~Pan, Dian Wang, Dong Yan, Fan Yang, Fei Deng, Feng Wang, Feng Liu, Guangwei Ai, Guosheng Dong, Hai Zhao, Hang Xu, Hao-Lun Sun, Hongda Zhang, Hui Liu, Jiaming Ji, Jian Xie, Juntao Dai, Kuncheng Fang, Lei Su, Liang Song, Lifeng Liu, Liyun Ru, Luyao Ma, Mang Wang, Mickel Liu, MingAn Lin, Nuolan Nie, Pei Guo, Ruiyang Sun, Zhang Tao, Tianpeng Li, Tianyu Li, Wei Cheng, Weipeng Chen, Xiangrong Zeng, Xiaochuan Wang, Xiaoxi Chen, Xin Men, Xin Yu, Xuehai Pan, Yan-Bin Shen, Yiding Wang, Yiyu Li, Youxin Jiang, Yuchen Gao, Yupeng Zhang, Zenan Zhou, and Zhiying Wu. 2023.
\newblock \href {https://api.semanticscholar.org/CorpusID:261951743} {Baichuan 2: Open large-scale language models}.
\newblock \emph{ArXiv}, abs/2309.10305.

\bibitem[{Yu et~al.(2023)Yu, Zhuang, Zhang, Meng, Ratner, Krishna, Shen, and Zhang}]{yu2023large}
Yue Yu, Yuchen Zhuang, Jieyu Zhang, Yu~Meng, Alexander Ratner, Ranjay Krishna, Jiaming Shen, and Chao Zhang. 2023.
\newblock Large language model as attributed training data generator: A tale of diversity and bias.
\newblock \emph{arXiv preprint arXiv:2306.15895}.

\bibitem[{Yuan et~al.(2023)Yuan, Jiao, Wang, Huang, He, Shi, and Tu}]{yuan2023gpt}
Youliang Yuan, Wenxiang Jiao, Wenxuan Wang, Jen-tse Huang, Pinjia He, Shuming Shi, and Zhaopeng Tu. 2023.
\newblock Gpt-4 is too smart to be safe: Stealthy chat with llms via cipher.
\newblock \emph{arXiv preprint arXiv:2308.06463}.

\bibitem[{Zefeng~Du(2023)}]{du-etal-2022-chinese-llama-2}
Longyue~Wang Zefeng~Du, Minghao~Wu. 2023.
\newblock Chinese-llama-2.
\newblock \url{https://github.com/longyuewangdcu/Chinese-Llama-2}.

\bibitem[{Zeng(2023)}]{zeng2023measuring}
Hui Zeng. 2023.
\newblock Measuring massive multitask chinese understanding.
\newblock \emph{arXiv preprint arXiv:2304.12986}.

\bibitem[{Zhao et~al.(2023{\natexlab{a}})Zhao, Zhou, Li, Tang, Wang, Hou, Min, Zhang, Zhang, Dong et~al.}]{zhao2023survey}
Wayne~Xin Zhao, Kun Zhou, Junyi Li, Tianyi Tang, Xiaolei Wang, Yupeng Hou, Yingqian Min, Beichen Zhang, Junjie Zhang, Zican Dong, et~al. 2023{\natexlab{a}}.
\newblock A survey of large language models.
\newblock \emph{arXiv preprint arXiv:2303.18223}.

\bibitem[{Zhao et~al.(2023{\natexlab{b}})Zhao, Zhao, Lu, Wang, Tong, and Qin}]{zhao2023chatgpt}
Weixiang Zhao, Yanyan Zhao, Xin Lu, Shilong Wang, Yanpeng Tong, and Bing Qin. 2023{\natexlab{b}}.
\newblock Is chatgpt equipped with emotional dialogue capabilities?
\newblock \emph{arXiv preprint arXiv:2304.09582}.

\bibitem[{Zheng et~al.(2023)Zheng, Chiang, Sheng, Zhuang, Wu, Zhuang, Lin, Li, Li, Xing et~al.}]{zheng2023judging}
Lianmin Zheng, Wei-Lin Chiang, Ying Sheng, Siyuan Zhuang, Zhanghao Wu, Yonghao Zhuang, Zi~Lin, Zhuohan Li, Dacheng Li, Eric Xing, et~al. 2023.
\newblock Judging llm-as-a-judge with mt-bench and chatbot arena.
\newblock \emph{arXiv preprint arXiv:2306.05685}.

\bibitem[{Zhong et~al.(2023)Zhong, Cui, Guo, Liang, Lu, Wang, Saied, Chen, and Duan}]{zhong2023agieval}
Wanjun Zhong, Ruixiang Cui, Yiduo Guo, Yaobo Liang, Shuai Lu, Yanlin Wang, Amin Saied, Weizhu Chen, and Nan Duan. 2023.
\newblock Agieval: A human-centric benchmark for evaluating foundation models.
\newblock \emph{arXiv preprint arXiv:2304.06364}.

\bibitem[{Zhou et~al.(2021)Zhou, Gopalakrishnan, Hedayatnia, Kim, Pujara, Ren, Liu, and Hakkani-Tur}]{zhou2021commonsense}
Pei Zhou, Karthik Gopalakrishnan, Behnam Hedayatnia, Seokhwan Kim, Jay Pujara, Xiang Ren, Yang Liu, and Dilek Hakkani-Tur. 2021.
\newblock Commonsense-focused dialogues for response generation: An empirical study.
\newblock In \emph{Proceedings of the 22nd Annual Meeting of the Special Interest Group on Discourse and Dialogue}, pages 121--132.

\bibitem[{Zhou et~al.(2022)Zhou, Muresanu, Han, Paster, Pitis, Chan, and Ba}]{zhou2022large}
Yongchao Zhou, Andrei~Ioan Muresanu, Ziwen Han, Keiran Paster, Silviu Pitis, Harris Chan, and Jimmy Ba. 2022.
\newblock Large language models are human-level prompt engineers.
\newblock \emph{arXiv preprint arXiv:2211.01910}.

\end{thebibliography}

\clearpage
\appendix

\begin{table*}[ht]
\centering
\small 
\resizebox{\textwidth}{!}{
\begin{tabularx}{\textwidth}
{>{\raggedright\arraybackslash}p{4cm} >{\raggedright\arraybackslash}p{11cm}}
\toprule 
\textbf{Task}    & \textbf{Definitions} \\
\midrule 
Knowledge-grounded Response Generation     & \textbf{Knowledge-grounded response generation} is a task of generating an informative response based on both dialogue context and the given external knowledge~\cite{santhanam2021rome}.   \\
Intent Classification    & \textbf{Intent classification} is a task of identifying which action the user wishes to take based on the dialogue context~\cite{louvan2020recent}.  \\
Slot Filling     & \textbf{Slot filling} is a task that maps the input slot key to the corresponding slot value based on the given dialogue context~\cite{chen2017survey}. 
 \\
Emotion Detection     & \textbf{Emotion detection} is a task of classifying the emotion of a speaker on a specific event in a dialogue~\cite{acheampong2020text}. \\
Personality-grounded Response Generation    &  \textbf{Personality-grounded response generation} is a task that generates an appropriate response that is consistent with the personality characteristics of the dialogue context~\cite{ma2020survey}.
 \\
Multi-turn Response Generation     & \textbf{Multi-turn response generation} is a task of generating a coherent response given a dialogue context~\cite{li2017dailydialog}. 
 \\
Dialogue Summarization     & \textbf{Dialogue summarization} is the process of extracting, summarizing, or refining key information from a multi-turn dialogue, turning it into a summary paragraph that can be used to present the main points of that multi-turn dialogue\cite{fengsurvey}. \\
Commonsense-aware Response Generation    & \textbf{Commonsense-aware response generation} is a task of generating an appropriate response incorporating correct commonsense knowledge~\cite{zhou2021commonsense}. \\
Dialogue Infilling     & \textbf{Dialogue infilling} is a task of infilling the missing utterance of the given dialogue that is consistent with the preceding and subsequent context~\cite{xue2022building}.  \\
Offensive Detection     & \textbf{Offensive detection} is a task that detects whether utterances contain uncivil, discriminatory, or aggressive content in the given dialogue~\cite{dinan2019build}.  \\
Dialogue Natural Language Inference    & \textbf{Dialogue natural language inference} is a task of inferring the semantic relationship between a certain part of a dialogue and a given hypothesis, including entailment, contradiction, and neutral~\cite{welleck2019dialogue}. \\
Relation Classification     & \textbf{Relations classification} is a task of inferring the interlocutor's interpersonal relationship from the information implied in the dialogue~\cite{jia2021ddrel}.
 \\
\bottomrule 
\end{tabularx}}
\caption{The definitions of all selected dialogue tasks.}
\label{tab:task_selection}
\end{table*}

\section{Task Selection}
\label{appendix:task_selection}

\subsection{Selection Process}
\label{appendis:selection_process}
We apply evaluation dimensions, including \emph{coherence, correctness, consistency} and \emph{safety} as guides and elaborately select tasks that focus on the corresponding evaluation dimension. Accordingly, those abilities can be reflected by the quality of task completion. The detailed selection process is provided as follows:
\begin{itemize}
    \item For coherence, We elect two tasks that are often focused on in the dialogue field, \emph{dialogue infilling}~\citep{xue2022building} and \emph{multi-turn dialog generation}~\citep{li2017dailydialog}.
    \item For correctness, we follow~\citet{zhao2023survey} and mainly examine the correctness of two aspects, including closed-scenario and open-scenario correctness.
    The closed-scenario correctness requires LLMs to generate the output only based on the given dialogue context or background knowledge.
    To this end, we select representative \emph{slot filling}~\citep{chen2017survey}, \emph{intent classification}~\citep{louvan2020recent}, along with \emph{knowledge-grounded response generation}~\citep{santhanam2021rome}.
    
    Conversely, the open-scenario correctness provides a testbed for probing the knowledge encoded by LLMs.
    We mainly select the ability to use commonsense correctly which is necessary for human-like dialogue systems, i.e., \emph{commonsense-aware response generation}~\citep{zhou2021commonsense}.
    \item For consistency, it mainly falls into two dimensions, including personalization consistency and semantic consistency~\citep{huang2020challenges}.
    For personalization consistency, we focus on capabilities necessary for real-human interactions, containing emotional perception, personality following, and relationship maintaining between speakers.
    As a result, we prioritize \emph{emotion detection}~\citep{acheampong2020text}, \emph{relation classification}~\citep{jia2021ddrel} and \emph{personality-grounded response generation}~\citep{ma2020survey} respectively.
    Semantic consistency refers to the actual semantic content contained in the dialogue context can entail the semantic content understood by humans, facilitating consistent response generation.
    Thus, we select the corresponding tasks, \emph{dialogue summarization}~\citep{fengsurvey} and \emph{dialogue NLI}~\citep{welleck2019dialogue}.
    \item For safety, some researchers~\citep{chang2023survey,bubeck2023sparks} have highlighted clear limitations of GPT-4, including toxic content generation.
    Therefore, we currently prioritize an important task that GPT-4 experts in, i.e., \emph{offensive detection}~\citep{dinan2019build}.
\end{itemize}

Consequently, we tease out $12$ dialogue tasks. The overall selection results are shown in Figure~\ref{fig:task_selection}.
Please see Appendix~\ref{appendix:task_definitions} for detailed task definitions.

\subsection{Task definitions}
\label{appendix:task_definitions}
The detailed task definitions are shown in Table~\ref{tab:task_selection}.

\section{Prompt Formatting}
\label{appendix:prompt_formatting}

Firstly, we clarify the core content of our prompt according to four key ingredients.
The key ingredients depict the functionality of a prompt for eliciting the abilities of GPT-4 to complete the goal, including goal description, input data, contextual information, and prompt style. 
\begin{itemize}
    \item \textbf{Goal description.} The goal description is typically a specific instruction that GPT-4 is expected to follow. 
    For a given dialogue task, we design the following information in natural language to describe the goal, including the background introduction and the generative step of the evaluation questions.
    By providing a well-clarified goal description, GPT-4 can more effectively understand the goal and generate the desired output.
    \item \textbf{Input data.} The input data provides the necessary information to guide the output generation that meets the requirements. The requirements primarily involve the difficulty of the evaluation instance.
    Inspired by human-level qualification exams, we heuristically set up the construction techniques for candidate options, formatted by the exact string, to control the difficulty.
    The clear and detailed input data allows GPT-4 to produce more controllable evaluation instances.
    \item \textbf{Contextual information.} In addition, contextual information is also essential to make prompts clear.
    In our creation, we find that it is necessary to provide some contextual information for explaining specific concepts appearing in the designed prompt.
    Therefore, we introduce the definition of multi-turn dialogue and the description of the dialogue task specifically to better depict our goal. 
    \item  \textbf{Prompt style.} A suitable prompt style can decompose the difficult task into several detailed sub-tasks to help GPT-4 accomplish the goal step by step.
    Inspired by this, we introduce the chain-of-thought (CoT) technique~\citep{wei2022chain}, which guides GPT-4 to generate evaluation instances step by step according to the order of the dialogue context, the task question, the candidate options, the problem-solving analysis, and the answer. 

\end{itemize}

When constructing each content of the four key ingredients, We mainly refer to the following design principles: (i) expressing the goal clearly, (ii) decomposing into easy, detailed sub-tasks, and (iii) utilizing a model-friendly format.
These design principles help create prompts that are clearer and easier to understand.
The final prompt is the exact string that concatenates each content of the four ingredients, as shown in Figure~\ref{fig:prompt_formatting}. In addition, the prompts of data generation for all tasks are listed in Table~\ref{tab:ic_prompt}-\ref{tab:ds_prompt}.

\section{Domain Bias}
\label{app:domain_bias}

The detailed descriptions of each domain are shown in Table~\ref{tab:domain_description}.
The selected domain involves two categories: daily life and professional knowledge.
Daily life mainly covers gourmet cooking, travel, household chores, film, neighborhood, workplace, music, shopping, games, and sports; while professional knowledge covers history, philosophy, sociology, psychology, economics, geography, physics, biology, computer science, and medicine.
The detailed descriptions we give are the specific topics that are typically talked about in each domain.

\section{Experimental Setup}
\subsection{LLMs to evaluate}
\label{appendix:llm_to_evaluate}
Table~\ref{tab:model_details} shows the details of pre-trained or supervised instruction-tuning LLMs for evaluation.

\begin{table*}[t!]
\centering
\footnotesize
\begin{tabular}{@{}llccc@{}}
    \toprule
    \textbf{Type} & \textbf{Model}   & \textbf{Parameters}   & \textbf{Access} & \textbf{Creator}  \\ \midrule
    \multirow{13}{*}{Pre-trained}   &\textrm{Baichuan2-13B}~\cite{Yang2023Baichuan2O}  & 13B     & Open  &  Baichuan  \\
    &Qwen-7B~\cite{Bai2023QwenTR}      & 7B       & Open  & Alibaba Cloud       \\
    &\textrm{InternLM-7B}~\cite{2023internlm}  & 7B       & Open &  Shanghai AI Laboratory \& SenseTime                 \\
     &\textrm{LLaMA2-70B}~\cite{Touvron2023Llama2O}    & 70B              & Open &  Meta         \\ 
     &\textrm{Mistral-7B}~\cite{jiang2023mistral}    & 7B            & Open  & Mistral AI \\ 
      &Baichuan-13B~\cite{Yang2023Baichuan2O} & 13B & Open  & Baichuan            \\
      &LLaMA2-13B~\cite{Touvron2023Llama2O} & 13B & Open  &   Meta        \\
       &\textrm{MOSS-Moon-003-Base}~\cite{sun2023moss}    & 16B          & Open & Fudan        \\
       &\textrm{Chinese LLaMA2-13B}~\cite{du-etal-2022-chinese-llama-2}   & 13B        & Open  &   Du et al.           \\
    &\textrm{LLaMA-65B}~\cite{Touvron2023LLaMAOA}   & 65B        & Open &  Meta                                   \\ 
    &LLaMA-13B~\cite{Touvron2023LLaMAOA}     & 13B           & Open  &  Meta     \\
    & LLaMA-7B~\cite{Touvron2023LLaMAOA}     & 7B           & Open  &  Meta    \\
    &\textrm{Falcon-7B}~\cite{falcon40b}     & 7B         & Open &  TII            \\
    
      \midrule
    
    \multirow{14}{*}{\shortstack[l]{Supervised \\ Instruction-tuning}}  & \textrm{GPT-4}~\cite{OpenAI2023GPT4TR} & \textit{undisclosed}   & API & OpenAI                  \\ 
    &\textrm{ChatGPT}~\cite{chatgpt}   & \textit{undisclosed}       & API &  OpenAI               \\
        &\textrm{Baichuan2-13B-Chat}~\cite{Yang2023Baichuan2O}   & 13B              & Open & Baichuan            \\
    &\textrm{InternLM-Chat-7B}~\cite{2023internlm} & 7B   & Open & Shanghai AI Laboratory \& SenseTime             \\
    &\textrm{Qwen-7B-Chat}~\cite{Bai2023QwenTR}  & 7B          & Open &  Alibaba Cloud         \\
    &\textrm{ChatGLM2-6B}~\cite{du2022glm}     & 6B          & Open &    Tsinghua \& Zhipu.AI       \\ 
    &\textrm{Mistral-7B-Instruct}~\cite{jiang2023mistral}    & 7B            & Open  & Mistral AI \\ 
    &\textrm{Baichuan-13B-Chat}~\cite{Yang2023Baichuan2O}    & 13B      & Open & Baichuan            \\
    &\textrm{Vicuna-13B}~\cite{zheng2023judging}     & 13B    & Open & LMSYS         \\
    &\textrm{Chinese Alpaca2-13B}~\cite{Cui2023EfficientAE}   & 13B     & Open &  Cui et al.              \\
    &\textrm{LLaMA2-7B-Chat}~\cite{Touvron2023Llama2O}   & 7B     & Open & Meta         \\
    &\textrm{MOSS-Moon-003-SFT}~\cite{sun2023moss}   & 16B                & Open & Fudan   \\ 
    &\textrm{Xwin-LM-7B}~\cite{xwin-lm} & 7B  & Open &  Xwin-LM Team     \\ 
    \bottomrule
    
\end{tabular}
\caption{The details of pre-trained or supervised instruction-tuning models LLMs for evaluation.}
\label{tab:model_details}
\end{table*}

\subsection{Evaluation Prompt Setup}
\label{appendix:prompt_setup}
\begin{figure*}
\centering
\includegraphics[width=1\textwidth,keepaspectratio]{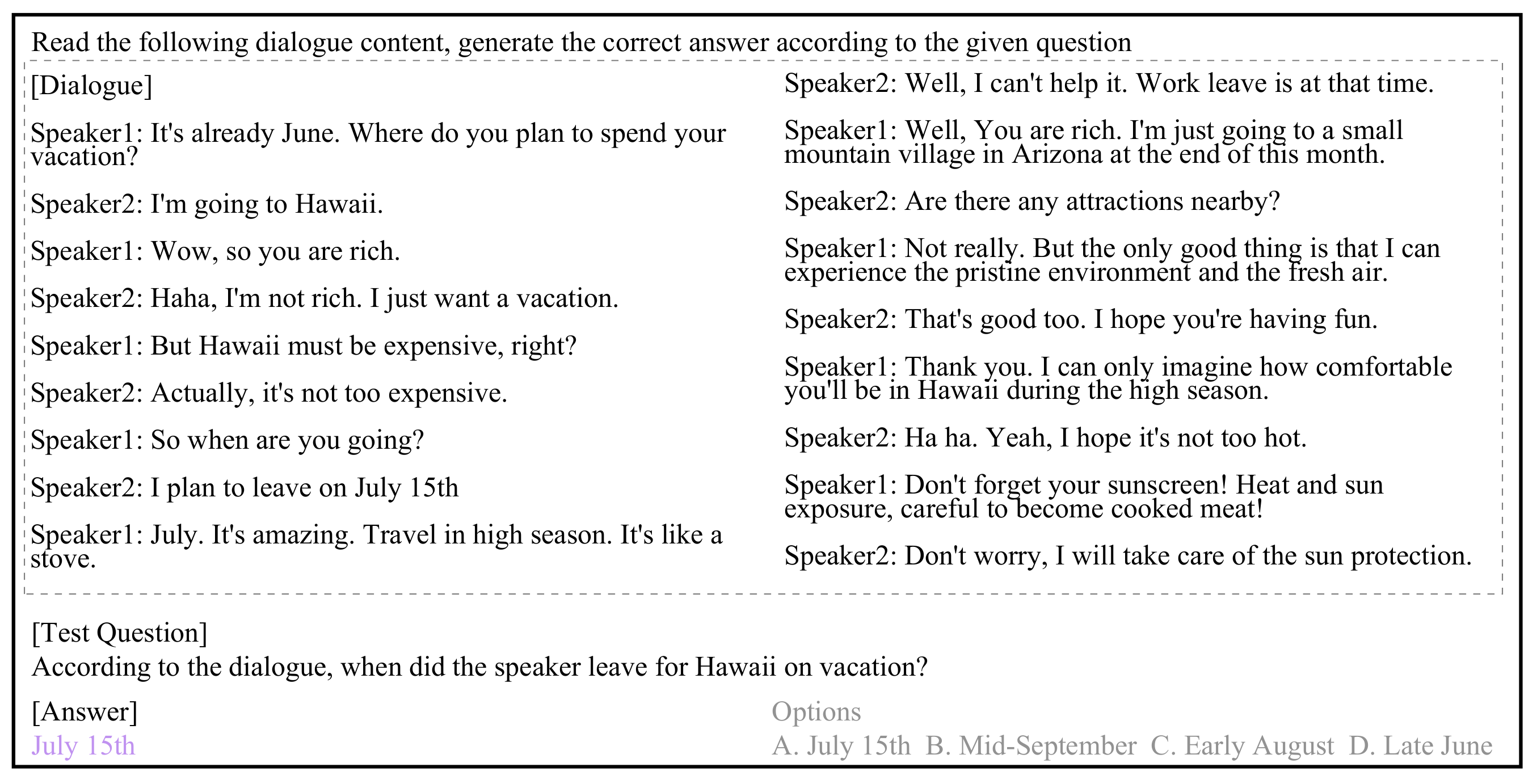}
\caption{An evaluation prompt for pre-trained LLMs is an exact string that concatenates all contents by ``\emph{Read the following dialogue content, generate the correct answer according to the given question[Dialogue]\{dialogue\_content\}[Test Question]\{test\_question\}[Answer]\{answer\_content\}}''. We take \emph{slot filling} task as an example. 
The \textcolor{mypurple}{purple} text is the answer content that LLMs need to calculate probability, which is selected from the \textcolor{mygrey}{Options} and calculated one by one.}
\label{fig:question_sample_pretrain}
\end{figure*}

\begin{figure*}
\centering
\includegraphics[width=1\textwidth,keepaspectratio]{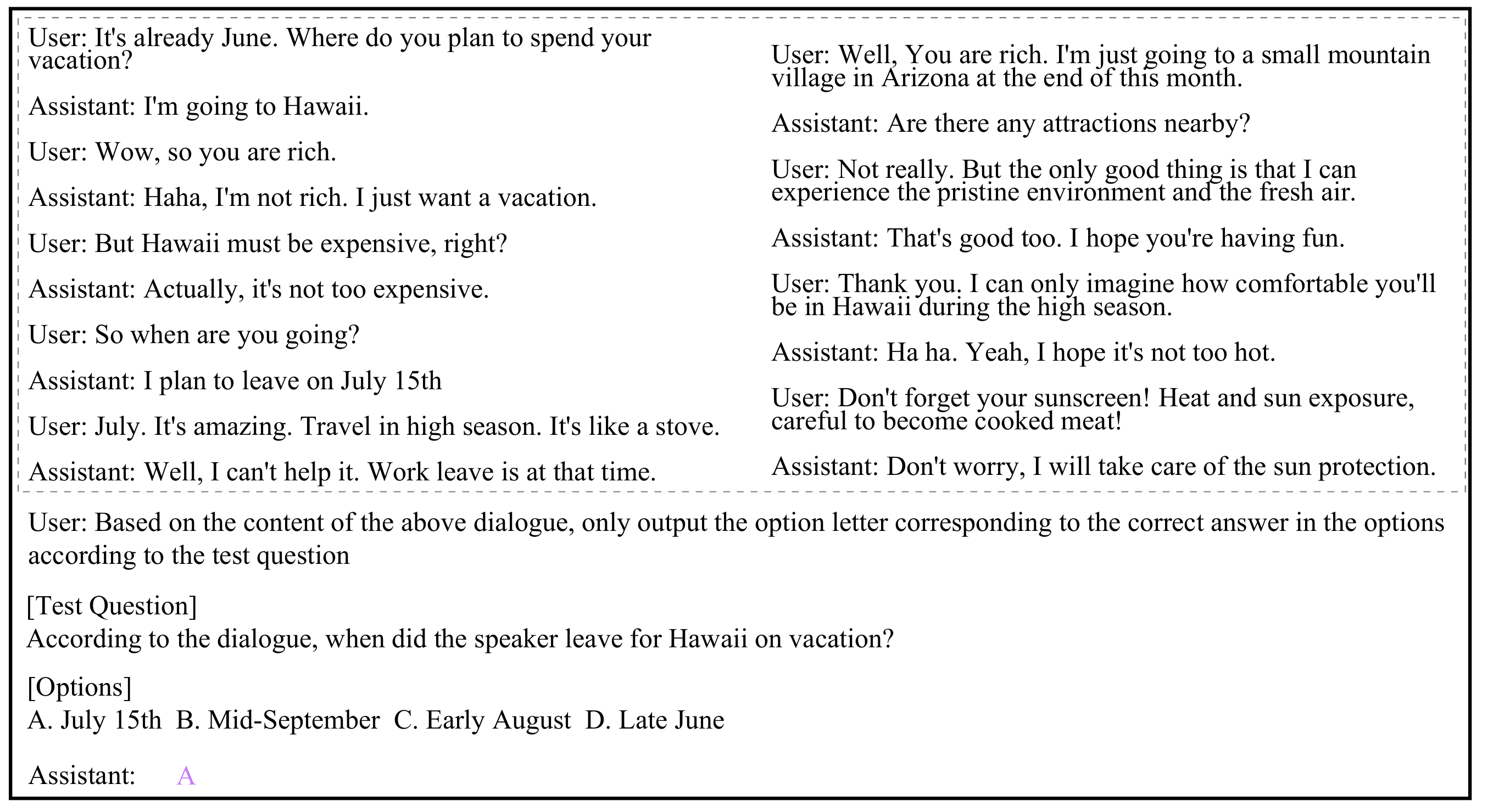}
\caption{An evaluation prompt for supervised instruction-tuning LLMs is an exact string, i.e., ``\emph{Based on the content of the above dialogue, only output the option letter corresponding to the correct answer in the candidate options according to the test question.[Test Question]\{test\_question\}[Options]\{option\_str\}}''. 
We regard the given dialogue as the history of chatting that has occurred between the user and the LLM. 
In the current interaction turn, this evaluation prompt is regarded as the user's question to the LLM, and then the LLM gives the option label.
We take \emph{slot filling} as an example.
The \textcolor{mypurple}{purple} text is the content generated by LLMs.}
\label{fig:question_sample_sft}
\end{figure*}

We evaluate LLMs in answer-only and zero-shot settings. Prompts used for two types of LLMs are shown in Figure~\ref{fig:question_sample_pretrain} and Figure~\ref{fig:question_sample_sft} respectively. 
Similar to the evaluation method, we use different instructions to induce LLMs to generate answers. 

\section{Main Results}
\label{appendix:main_results}
A central objective for our evaluation is to achieve a common and unified understanding of the corresponding capabilities of LLMs as human-like dialogue systems.
We first evaluate the pre-trained LLMs using DialogBench to provide a baseline of LLMs' capabilities as human-like dialogue systems.
Further, we evaluate the supervised instruction-tuning LLMs and analyze the impact of instruction fine-tuning on LLMs as human-like dialogue systems.

\subsection{Pretrained LLMs}
Overall and task-specific scores in Chinese and English DialogBench are reported at the top of Table~\ref{tab:main_result} and~\ref{tab:main_result_en} respectively. 
The overall score of all LLMs on English DialogBench is slightly better than the score on Chinese DialogBench.
Additionally, the overall performance of all LLMs on each task generally has the same trend on Chinese and English DialogBench.
Next, we mainly conduct analysis based on the results on Chinese DialogBench.
We first give an overall analysis and further highlight findings at the task level from evaluation dimension perspectives.
\paragraph{Overall Analysis.}
On this challenging benchmark, surprisingly we discover that some pre-trained LLMs have pretty good performances.
Specifically, Baichuan2-13B presents the best performance, scoring an overall accuracy of $54.43\%$ on DialogBench.
Qwen-7B and InternLM-7B follow closely behind with overall accuracy scores of $53.63\%$  and $53.37\%$ respectively. 
For other pre-trained LLMs, despite their relatively poorer performance, most of them can score above $43\%$.
Overall, there is still much room for improvement in these capabilities for pre-trained LLMs as human-like dialogue systems.
We further observe that LLaMA-13B has higher overall accuracy than LLaMA-7B (e.g., $46.23\%$ vs. $44.17\%$), and correspondingly LLaMA-65B has higher overall accuracy than LLaMA-13B (e.g., $47.72\%$ vs. $46.23\%$).
It suggests that the model scale is monotonically correlated with the model accuracy win rate within a model family.

\paragraph{Dimension-specific Analysis.}
For the $4$ tasks in the correctness dimension, the average accuracy scores of all pre-trained LLMs are $72.63\%$ for slot filling (SF), $61.61\%$ for knowledge-grounded response generation (KRG), $54.55\%$ for intent classification (IC), and $51.81\%$ for commonsense-aware response generation (CRG).
Accordingly, the best scores are $81.84\%$, $67.93\%$, $61.79\%$, and $60.57\%$ respectively.
As these suggest, most pre-trained LLMs can perform well on slot filling but have relatively poor performance on the other $3$ tasks about correctness.
Furthermore, we observe that the margin varies across different tasks: the largest margin is on knowledge-grounded response generation (KRG) where LLaMA2-70B achieves an accuracy of $67.93\%$ compared to second place from Baichuan2-13B at $63.18\%$, whereas the smallest margin is for slot filing (SF), i.e., $81.84\%$ for LLaMA2-70B vs. $80.91\%$ for QWen-7B.
In general, the margins between the various pre-trained LLMs are not significant, which indicates that these pre-trained LLMs have modestly different performances in correctness-related abilities.

For the $5$ tasks in the consistency dimension, we divide these tasks into two groups for analysis, including personalization consistency and semantic consistency.
For the $3$ tasks about personalization consistency, the average accuracy scores of all LLMs are $55.09\%$ for emotion detection (ED), $29.97\%$ for relation classification (RC), and $26.46\%$ for personality-grounded response generation (PRG).
The best scores are $69.93\%$, $56.89\%$, and $28.90\%$ respectively.
These show that pre-trained LLMs as a whole have good performances in emotion perception, whereas the performance on personality following is unsatisfactory.
We speculate that the pre-trained LLMs have not seen many instances related to personality following. 
The finding about personality is consistent with~\cite{safdari2023personality}.
For the $2$ tasks about semantic consistency, the average accuracy scores of all LLMs are $59.45\%$ for dialogue summarization (DS) and $36.49\%$ for dialogue NLI (NLI). The best scores are $69.52\%$, and $45.11\%$ respectively.
The decent performance on dialogue summarization indicates that pre-trained LLMs perform well in maintaining semantic alignment, however, it is still relatively difficult to maintain semantic consistency that requires one-step reasoning, as shown by the performance on dialogue NLI.
Overall, we see significant heterogeneity across the results on consistency-related tasks, which may be because each task requires different levels of abilities.

For the $2$ tasks in the coherence dimension, the average accuracy scores of all LLMs are $44.01\%$ for dialogue infilling (DI) and $47.22\%$ for multi-turn response generation (MRG), along with the best scores of $47.56\%$ and $54.58\%$ respectively.
It indicates that the average performance of LLMs on both tasks is relatively similar, and there is still much room for improvement in maintaining dialogue coherency.
For offensive detection (OD) in the safety dimension, the average accuracy of all LLMs is $53.44\%$ and the best performance is an accuracy of $62.57\%$, which shows that most pre-trained LLMs have scores around $50\%$ and empower a certain capability of offensive detection.

Overall, current pre-trained LLMs perform relatively well on correctness-related tasks and have greater room for improvement on tasks related to coherence and safety. For consistency-related tasks, it is necessary for pre-trained LLMs to continue to be optimized to possess corresponding high-order capabilities.

\begin{table*}[t!]
\centering
\resizebox{\textwidth}{!}{
\begin{tabular}{@{}lccccccccccc@{}}
    \toprule
    \multirow{2}{*}{\textbf{Model}} & \multicolumn{11}{c}{\textbf{Daily Life}} \\ \cmidrule(l){2-12}
    & \shortstack{\textbf{Gourmet} \\ \textbf{Cooking}} & \textbf{Travel} & \shortstack{\textbf{Household} \\ \textbf{Chores}} & \textbf{Film} & \textbf{Neighborhood} & \textbf{Workplace} & \textbf{Musics} & \textbf{Shopping} & \textbf{Games} & \textbf{Sports} & \textbf{Avg.} \\ 
    \midrule
        GPT-4 & 77.80 & 83.11 & 74.60 & 84.49 & 79.63 & 83.67 & 76.84 & 78.63 & 79.18 & 80.68 & 79.86 \\ 
        ChatGPT & 67.75 & 67.58 & 60.80 & 68.54 & 66.08 & 76.51 & 67.22 & 67.61 & 67.40 & 69.65 & 67.91 \\ 
        Baichuan-2-13B-Chat & \underline{62.01} & \textbf{64.65} & \textbf{60.17} & \textbf{64.75} & \textbf{65.24} & \textbf{68.50}& \underline{59.65} & \textbf{66.20} & \textbf{64.77} & \underline{63.74} & \textbf{63.97} \\ 
        InternLM-Chat-7B & 60.66 & 61.16 & \underline{58.86} & \underline{60.72} & \underline{63.09} & \underline{67.04} & \textbf{60.49} & \underline{63.57} & 61.26 & \textbf{64.88} & \underline{62.17} \\
        Qwen-7B-Chat & 61.92 & \underline{61.19} & 57.95 & 59.93 & 62.03 & 64.89 & 57.37 & 63.46 & \underline{61.35} & 62.42 & 61.25 \\ 
        ChatGLM-6B & 55.28 & 59.60 & 55.70 & 52.71 & 57.23 & 57.33 & 52.02 & 56.40 & 56.72 & 58.06 & 56.11 \\ 
        Baichaun-13B-Chat & \textbf{62.11} & 55.40 & 57.59 & 49.50 & 60.35 & 56.97 & 50.50 & 56.01 & 52.77 & 56.13 & 55.73 \\ 
        Mistral-7B-Instruct-v0.2 & 54.04 & 54.34 & 52.95 & 51.70 & 55.86 & 55.19 & 53.20 & 52.71 & 53.56 & 53.54 & 53.71 \\
        Vicuna-13B & 45.13 & 41.51 & 40.08 & 40.28 & 38.67 & 44.72 & 37.19 & 43.41 & 36.96 & 41.44 & 40.94 \\ 
        Chinese Alpaca2-13B & 42.44 & 39.07 & 39.03 & 37.27 & 36.33 & 45.70 & 37.59 & 39.92 & 42.89 & 40.65 & 40.09 \\ 
        LLaMA2-Chat-7B & 39.34 & 38.96 & 38.61 & 35.27 & 38.28 & 38.30 & 37.12 & 38.18 & 36.96 & 35.38 & 37.64 \\ 
        MOSS-Moon-003-SFT & 31.68 & 30.35 & 33.76 & 29.06 & 29.88 & 29.36 & 29.70 & 32.75 & 30.83 & 33.87 & 31.12 \\ 
        Xwin-7B & 25.47 & 27.30 & 28.27 & 25.05 & 23.44 & 35.45 & 23.71 & 28.29 & 28.26 & 27.39 & 27.26 \\ 
        \midrule
        Avg. & 52.74 & 52.63 & 50.64 & 50.71 & 52.01 & 55.66 & 49.43 & 52.86 & 51.76 & 52.91 & 52.14 \\ 
        \bottomrule
        \toprule
        \multirow{2}{*}{\textbf{Model}} & \multicolumn{11}{c}{\textbf{Professional Knowledge}} \\ \cmidrule(l){2-12}
        ~ & \textbf{History} & \textbf{Philosophy} & \textbf{Sociology} & \textbf{Psychology} & \textbf{Economics} & \textbf{Geography} & \textbf{Physics} & \textbf{Biology} & \shortstack{\textbf{Computer} \\ \textbf{Science}} & \textbf{Medicine} & \textbf{Avg.} \\ 
        \midrule
        GPT-4 & 82.87 & 77.64 & 79.39 & 80.84 & 84.15 & 85.05 & 81.59 & 86.09 & 86.11 & 88.53 & 83.23 \\ 
        ChatGPT & 71.59 & 65.97 & 72.84 & 71.36 & 75.06 & 71.43 & 71.15 & 73.39 & 79.12 & 77.29 & 72.92 \\ 
        Baichuan-2-13B-Chat & \textbf{65.89} & \textbf{64.33} & \textbf{68.17} & \underline{69.12} & \textbf{71.08} & \textbf{66.36} & \textbf{65.46} & \textbf{72.34} & \underline{73.30} & \underline{73.22} & \textbf{68.93} \\ 
        InternLM-Chat-7B & 63.65 & 58.10 & 63.18 & \textbf{69.37} & 67.50 & 63.67 & \underline{62.92} & \underline{68.74} & \textbf{74.29} & \textbf{75.12} & \underline{66.65} \\ 
        Qwen-7B-Chat & \underline{64.73} & \underline{61.48} & \underline{65.19} & 66.03 & \underline{68.82} & \underline{64.72} & 61.70 & 66.45 & 68.47 & 71.09 & 65.87 \\ 
        ChatGLM-6B & 59.03 & 59.11 & 62.17 & 60.45 & 64.60 & 60.99 & 58.27 & 59.00 & 62.19 & 66.29 & 61.21 \\ 
        Baichaun-13B-Chat & 58.67 & 53.24 & 56.64 & 58.12 & 59.00 & 61.11 & 52.61 & 59.30 & 58.54 & 64.36 & 58.16 \\
        Mistral-7B-Instruct-v0.2 & 55.58 & 52.02 & 55.94 & 53.14 & 59.54 & 53.44 & 55.73 & 61.35 & 54.21 & 62.16 & 56.31 \\
        Vicuna-13B & 48.17 & 44.94 & 45.27 & 47.67 & 49.37 & 50.41 & 48.31 & 49.95 & 46.01 & 48.52 & 47.86 \\ 
        Chinese Alpaca2-13B & 39.84 & 42.11 & 45.27 & 42.19 & 47.88 & 45.78 & 46.17 & 46.39 & 48.29 & 47.96 & 45.19 \\ 
        LLaMA2-Chat-7B & 39.32 & 40.49 & 40.64 & 41.78 & 40.14 & 40.80 & 45.07 & 41.56 & 41.69 & 40.58 & 41.21 \\ 
        MOSS-Moon-003-SFT & 30.68 & 32.39 & 30.58 & 32.86 & 33.63 & 31.61 & 31.46 & 34.17 & 31.89 & 37.79 & 32.71 \\ 
        Xwin-7B & 29.44 & 25.51 & 29.98 & 30.63 & 28.76 & 30.31 & 30.79 & 25.74 & 23.23 & 32.04 & 28.64 \\ 
        \midrule
        Avg. & 54.57 & 52.10 & 55.02 & 55.66 & 57.66 & 55.82 & 54.71 & 57.27 & 57.49 & 60.38 & 56.07\\ 
    \bottomrule
\end{tabular}}
\caption{Accuracy of supervised instruction-tuning LLMs in Chinese DialogBench for all $20$ domains. All domains are mainly divided into two categories, including daily life and professional knowledge.
\textbf{Bold} and \underline{underlined} indicate the best results and the second-best results respectively except for GPT-4 and ChatGPT.}
\label{tab:domain_result}
\end{table*}

\subsection{Supervised Instruction-tuning LLMs}
The overall and task-specific scores in Chinese and English DialogBench are reported at the bottom of Table~\ref{tab:main_result} and~\ref{tab:main_result_en}.
We also conduct an analysis based on the results on Chinese DialogBench, along with giving an overall analysis and task-level analysis respectively.
Additionally, we analyze the performance changes of pre-trained LLMs and instruction-tuning LLMs in the same model family on different tasks.

\paragraph{Overall Analysis.}
As shown in Table~\ref{tab:main_result}, the results show that the overall performances of different models are different, and the performance of the same LLM on different dialogue tasks also varies widely.
We further observe that: 
(1) GPT-4 presents the best performance on overall accuracy with $81.54\%$, which basically represents the best performance that existing supervised instruction-tuning LLMs can achieve.
This excellent score also indicates that GPT-4 has strong capabilities as a human-like dialogue system.
In addition, ChatGPT achieves an overall scores of $70.42\%$, ranking second.
(2)They are closely followed by Baichuan2-13B-Chat with $66.44\%$, and InterLM-Chat-7B with $64.40\%$. The performance gap between GPT-4 and the best open-source LLMs ($81.54\%$ vs. $66.44\%$) shows that there is still much room for improvement in the capabilities that LLMs should have as human-like dialogue systems.
Compared with ChatGPT, Baichuan2-13B-Chat have achieved better performances on $3$ out of $12$ tasks, 
which indicates that Baichuan2-13B-Chat currently has pretty good capabilities related to human likeness.
(3) The instruction-tuning LLMs achieve higher scores than the corresponding pre-trained LLMs on most dialogue tasks (e.g., QWen-7B vs. QWen-7B-Chat, Baichuan2-13B-Base vs. Baichuan2-13B-Chat), which suggests that instruction tuning is an efficient and effective means for improving the capabilities that LLMs should have as human-like dialogue systems.

\paragraph{Dimension-specific Analysis.}
For the $4$ tasks in the correctness dimension, GPT-4 and ChatGPT achieve scores of over $81.46\%$ and $76.69\%$ on all tasks, including slot filling (SF), intent classification (IC), knowledge-grounded response generation (KRG), and commonsense-aware response generation (CRG), which demonstrates that it is not unachievable currently to improve the correctness-related capabilities of LLMs.
Most other LLMs (e.g., Qwen-7B, ChatGLM, Baichuan variants, InternLM) also have impressive results on these tasks, with the average accuracy remaining around $73.13\%$, $70.19\%$, $74.07\%$, and $66.31\%$ respectively. 
This shows that most supervised instruction-tuning LLMs can understand the intent and slot in the dialogue context, along with selecting appropriate knowledge or commonsense for generating responses with reasonable accuracy.
In addition, the supervised instruction-tuning LLMs achieve higher scores than the pre-trained LLMs in the same model family (e.g., QWen, Baichuan2, and InternLM) on almost all corresponding tasks, which indicates that instruction finetuning benefits LLMs improving those capabilities related to correctness.
However, there is no such improvement in slot filling (SF), probably because this task is simple enough and the pre-trained LLMs already have quite good capabilities.

For the $5$ tasks in the consistency dimension, we also analyze personalization and semantic perspectives respectively.
For the $3$ tasks about personalization consistency, GPT-4 only achieves $61.90\%$ and $72.83\%$ in emotion detection (ED) and personality-grounded response generation (PRG), and ChatGPT obtains relatively inferior scores correspondingly.
Most other LLMs also perform unsatisfactorily on these two tasks, with the average scores of the remaining LLMs around $39.19\%$ and $34.96\%$.
We speculate that the current positioning of LLMs is assistant AI, which would weaken the LLMs' abilities of emotional perception and personality following.
Relatively speaking, LLMs perform relatively better on relation classification (RC), with an average score of all LLMs except GPT-4 and ChatGPT around $40.93\%$. 
But overall, all LLMs still have much room for improvement in tasks related to personalization consistency.
Interestingly, most supervised instruction-tuning LLMs achieve inferior scores on emotion classification than the pre-trained LLMs in the same model family, such as QWen-7B, Baichuan2-13B and InternLM-7B.
It might be due to the fact that the positioning of assistant AI enables instruction tuning to focus on the ability to complete tasks, abandoning the ability to perceive emotions.
For the $2$ tasks about semantic consistency, there is the same conclusion that LLMs perform well on dialogue summarization (DS), e.g., GPT-4 with a score of $90.22\%$, but perform relatively poorly on dialogue NLI (NLI) that requires one-step reasoning (e.g., GPT-4 with a score of $75.39\%$).
In addition, we observe that instruction tuning can also generally improve consistency-related capabilities.

For the $2$ tasks in the coherence dimension, GPT-4 achieves scores of $79.22\%$ and $77.75\%$ on dialogue infilling (DI) and multi-turn response generation (MRG) respectively. 
Accordingly, ChatGPT achieves scores of $65.53\%$ and $70.72\%$ respectively.
The other LLMs have relatively inferior performances, with the average accuracy of $48.20\%$ and $50.35\%$.
Furthermore, instruction tuning also improves coherence-related capabilities compared with the pre-trained LLMs and the supervised instruction-tuning LLMs in the same model family.
These indicate that although instruction tuning has enhanced the LLMs' ability to generate coherent responses to a certain extent, there is still much room for improvement.
For offensive detection (OD) in the safety dimension, GPT-4 and ChatGPT obtain scores of $83.15\%$ and $61.50\%$ respectively, which suggest that there is still room for research on this task.
In addition, Some top LLMs (e.g., Baichuan2-13B and QWen-7B) achieve higher scores via instruction tuning, however, this improvement does not appear on other LLMs.

Overall, there is the same trend of performances on different evaluation dimensions for supervised instruction-tuning LLMs as pre-trained LLMs.
The difference is that supervised instruction-tuning LLMs generally have stronger performances than the corresponding pre-trained LLMs.

\begin{table*}[t]
\centering
\footnotesize
\resizebox{\textwidth}{!}{
\begin{tabularx}{\textwidth}
{>{\raggedright\arraybackslash}p{1.5cm} >{\centering\arraybackslash}p{1.5cm} >{\raggedright\arraybackslash}p{11.7cm}}
    \toprule
        & \textbf{Doamin} & \textbf{Description}    \\ \midrule
    \multirow{10}{*}[-2.5cm]{Daily Life} 
& Gourmet Cooking & Recipes, cooking techniques, ingredients, food culture, kitchen gadgets, famous chefs, restaurant reviews, food blogs or channels, dietary preferences, and international cuisines. \\
& Travel & Travel destinations, travel experiences, cultural differences, local cuisine, transportation modes, packing and preparation, budget traveling, travel tips, accommodations, travel photography, must-see landmarks. \\
& Household Chores & Household chores, family relationships, parenting, home organization, cooking and meal planning, home improvement, family traditions, family budgeting, child-rearing methods, family vacations. \\ 
& Film & Movie genres, TV shows, favorite actors and actresses, film directors, streaming platforms, movie or TV show reviews, upcoming releases, movie soundtracks, cinematography, film festivals, industry news and trends. \\
& Neighborhood & Neighbor interactions, community events, neighborhood safety, local issues and improvements, shared spaces and facilities, cultural differences, neighborhood history, local businesses and services, how to be a good neighbor. \\ 
& Workplace & Workplace environment, company culture, career goals, job search, interview experiences, promotions, networking, work-life balance, job satisfaction, professionalism, office politics, leadership and management styles, time management, conflict resolution. \\
& Music & Musical genres, favorite artists, concerts and live performances, music history, learning to play instruments, singing, musical influences, songwriting, music streaming platforms and technology, music festivals, soundtracks or background music in movies and TV shows. \\
& Shopping & Shopping habits, fashion trends, sales and discounts, online shopping, favorite stores, product reviews, shopping tips, sustainable and ethical shopping, gift ideas, shopping experiences. \\
& Games & Video game genres, game consoles, online gaming, mobile gaming, esports, favorite games, game characters, gameplay strategies, gaming communities, game developers and publishers, virtual reality gaming, gaming nostalgia, and upcoming game releases. \\ 
& Sports & Sports history, sports rules and regulations, workout routines, sports nutrition, exercise benefits, sports careers, sports injuries and prevention, sports movies and documentaries. \\
    \midrule
    \multirow{10}{*}[-2.4cm]{\shortstack{Professional \\ Knowledge}} 
& History & Ancient civilizations, historical figures, famous battles and wars, historical landmarks, cultural and social history, myths and legends, historical discoveries and inventions, historical timelines and eras, historical art and literature. \\
& Philosophy & Existentialism, metaphysics, ethics, epistemology, logic, philosophy of mind, political philosophy, eastern philosophy, religious philosophy, philosophical debates and theories, famous philosophers. \\
& Sociology & Social inequality, race and ethnicity, gender roles, family dynamics, education, religion, crime and deviance, urbanization, globalization, social movements, media and communication, mental health and well-being. \\
& Psychology & Cognitive psychology, developmental psychology, social psychology, personality theories, mental disorders, therapeutic approaches, emotions and motivation, memory and learning, psychological research methods, famous psychologists. \\
& Economics & Economic theories, supply and demand, fiscal policies, monetary policies, international trade, financial markets, economic indicators, income inequality, globalization, economic development. \\
& Geography & Physical geography, human geography, geographic locations, climate and weather, cartography and mapping, natural resources, cultural and regional diversity, geopolitics, landforms and geology, and environmental issues. \\
& Physics & Classical mechanics, quantum mechanics, relativity theory, thermodynamics, electromagnetism, particle physics, astrophysics, cosmology, string theory, scientific breakthroughs and discoveries, famous physicists, experiments and observations, theoretical vs experimental physics. \\
& Biology & Cellular biology, genetics, evolution, ecology, animal behavior, anatomy and physiology, biodiversity, taxonomy, microbiology, biotechnology, conservation, plant biology, marine biology. \\
& Computer Science & Programming languages, algorithms, data structures, artificial intelligence, machine learning, cybersecurity, software development, hardware components, computer networking, operating systems, computer history, coding projects, technology trends, computer science education, career paths in technology. \\
& Medicine & Medical advancements, healthcare system, alternative medicine, medical ethics, research and discoveries, diseases and treatments, mental health, nutrition and diet, medical careers. \\
    \bottomrule
\end{tabularx}}
\caption{The specific topics that are typically talked about in each domain.}
\label{tab:domain_description}
\end{table*}

\begin{table*}[t]
\centering
\resizebox{\textwidth}{!}{
\begin{tabularx}{\textwidth}
{>{\raggedright\arraybackslash}p{16cm}}
    \toprule
    \textbf{Prompt for Intent Classification} \\
    \midrule
We are currently testing the annotation capabilities of data annotators for intent classification in multi-turn dialogues. Multi-turn dialogue refers to a chat record generated by two speakers engaging in continuous interactions using natural language. Intent classification refers to determining a specific view, plan, or action to be taken by a speaker from the information reflected in the multi-turn dialogue. 

~

You are an expert at intent classification in multi-turn dialogues, and now please act as the test creator. Firstly, please generate a two-party dialogue with at least 10 turns(10 turns=20 utterances), followed by a multi-choice question of intent understanding that requires a comprehensive understanding of the context to answer. The dialogue content should be relevant to \{Domain\}.

~

Before setting the question, you should randomly set the personality of both speakers, and you have a certain probability of setting the personality of the speakers to be unfriendly so as to generate unfriendly, ironic, offensive, quarreling, specifying, weird, and negative content. It simulates the dialogue scenes of real human beings and helps to better study the intent understanding ability under the various scenarios.

~

When you generate the dialogue, please avoid using interrogative sentences in the dialogue. The last turn of the dialogue is not the end of the dialogue session. The multi-choice question you generated should be as difficult as possible, requiring a comprehensive understanding of the whole dialogue rather than a certain turn to answer. Meanwhile, the information involved in candidate options should be from the dialogue, which makes the wrong options even more confusing. The question can be answered only based on the dialogue without external knowledge. Please especially note that the correct answer should not be a snippet of the dialogue, which makes the question too simple. All candidate options need to be similar in terms of length and content to make it more difficult to distinguish between correct and wrong candidates. All candidate options can be determined their correctness after comprehensive understanding and extensive reasoning on the dialogue. 

~

The format of the question is as follows: \{"id": "xx", "task": "Intent Classification", "domain": "\{Domain\}", "speaker1 personality": "xx", "speaker2 personality": "xx", "dialogue": [\{"speaker1": "xx", "speaker2": "xx"\}, \{"speaker1": "xx", "speaker2": "xx"\}, \{"speaker1": "xx", "speaker2": "xx"\}, ..., \{"speaker1": "xx", "speaker2": "xx"\}], "test\_question": "xx", "option": "xx", "analysis": "xx", "label": "xx"\}. Where "id" denotes a randomly generated id, "speaker1 personality" denotes the character personality of speaker1, "speaker2 personality" denotes the character personality of speaker2, "dialogue" denotes more than 10 turns of two-person English dialogue where the two speakers are represented by speaker1 and speaker2  in each turn of interaction. "test\_question" denotes the multi-choice question, "option" denotes a dictionary that contains 4 candidate options, with capital letter sequence identifiers as the key. "analysis" denotes the reason for choosing that correct option. "label" denotes the correct option represented by a capital letter. Note that the generated question should be in JSON format. Now, please set the question. \\
    \bottomrule
\end{tabularx}}
\caption{The prompt for data generation of intent classification by GPT-4.}
\label{tab:ic_prompt}
\end{table*}

\begin{table*}[t]
\centering
\resizebox{\textwidth}{!}{
\begin{tabularx}{\textwidth}
{>{\raggedright\arraybackslash}p{16cm}}
    \toprule
    \textbf{Prompt for Emotion Detection} \\
    \midrule
We are testing data annotators on their annotation capabilities for emotion classification multi-turn dialogues. A multi-turn dialogue is a chat transcript produced by multiple turns of sustained interaction between two speakers using natural language. Emotion detection refers to classifying the emotion of a speaker on a specific event from the information reflected in the multi-turn dialogue. 

~

You are an expert at emotion classification in multi-turn dialogues, and now please act as the test creator. You begin by assuming the emotions of both sides of the dialogue, with emotions limited to eight types: disgust, fear, disappointment, neutrality, anger, sadness, joy, and surprise. Then please generate a two-party dialogue with at least 10 turns(10 turns=20 utterances) reflecting the emotions of both sides, followed by a multi-choice question of emotion classification that requires a comprehensive understanding of the context to answer. The dialogue content should be relevant to \{Domain\}.

~

Before setting the question, you should randomly set the personality of both speakers, and you have a certain probability of setting the personality of the speakers to be unfriendly so as to generate unfriendly, ironic, offensive, quarreling, specifying, weird, and negative content. It simulates the dialogue scenes of real human beings and helps to better study the intent understanding ability under the various scenarios.

~

When you generate the dialogue, please avoid using interrogative sentences in the dialogue. The content of the last turn of the sub-dialogue is not a closing statement. Your questions must be as difficult as possible, and the dialogue content does not literally reflect any candidate options directly, but by analyzing the content or the tone of voice in the dialogue, etc., you can clearly determine which candidate emotions are correct.

~

The format of the question is as follows: \{"id": "xx", "task": "Emotion Detection", "speaker1 emotion": "xx", "speaker2 emotion": "xx", "domain": "\{Domain\}",  "speaker1 personality": "xx",  "speaker2 personality": "xx",  "dialogue": [\{"speaker1": "xx", "speaker2": "xx"\}, \{"speaker1": "xx", "speaker2": "xx"\}, \{"speaker1": "xx", "speaker2": "xx"\}, ..., \{"speaker1": "xx", "speaker2": "xx"\}], "test\_question": "xx", "option": "xx", "analysis": "xx", "label": "xx"\}. Where "id" denotes a randomly generated id, "speaker1 emotion" is the randomly assumed emotion of speaker1, "speaker2 emotion" is the randomly assumed emotion of speaker2,  "speaker1 personality" denotes the character personality of speaker1, "speaker2 personality" denotes the character personality of speaker2, "dialogue" denotes more than 10 turns of two-person English dialogue where the two speakers are represented by speaker1 and speaker2  in each turn of interaction. "test\_question" denotes the multi-choice question about a speaker's emotional attitude toward a certain event. "option" is a dictionary containing four candidate options, with capital letters representing the serial number identifier as the key. The candidate emotions are limited to eight types: disgust, fear, disappointment, neutrality, anger, sadness, joy, and surprise, of which only one type is correct. "analysis" denotes the reason for choosing the correct option. "label" denotes the correct option represented by a capital letter. Note that the generated question should be in JSON format. Now, please set the question. \\

    \bottomrule
\end{tabularx}}
\caption{The prompt for data generation of emotion detection by GPT-4.}
\end{table*}

\begin{table*}[t]
\centering
\resizebox{\textwidth}{!}{
\begin{tabularx}{\textwidth}
{>{\raggedright\arraybackslash}p{16cm}}
    \toprule
    \textbf{Prompt for Commonsense-aware Response Generation} \\
    \midrule
We are currently testing the annotation capabilities of data annotators for response selection in multi-turn commonsense-aware dialogues. Multi-turn dialogue refers to a chat record generated by two speakers engaging in continuous interactions using natural language. Commonsense, namely generic knowledge, refers to the basic knowledge that a mentally and physically grown-up adult should possess to live in society, including survival skills (self-care ability), basic labor skills, common knowledge in natural science, humanities and social science, etc. The commonsense-aware response selection in multi-turn dialogue refers to selecting the response by annotators according to the context as the response to the utterance by speaker1 in the last turn, which uses correct commonsenses.

~

You are an expert at commonsense-aware response selection in multi-turn dialogues, and now please act as the test creator. Firstly, please generate a two-party dialogue with at least 10 turns(10 turns = 20 utterances) and a piece of commonsense, followed by a multi-choice question based on this common sense. The dialogue content should be relevant to \{Domain\}.

~

Before setting the question, you should randomly set the personality of both speakers, and you have a certain probability of setting the personality of the speakers to be unfriendly so as to generate unfriendly, ironic, offensive, quarreling, specifying, weird, and negative content. It simulates the dialogue scenes of real human beings, and helps to better study the response selection ability in multi-turn commonsense-aware dialogues under the various scenarios.

~

When you generate the dialogue, please avoid using interrogative sentences in the dialogue. The last turn of the dialogue is not the end of the dialogue session. The last turn of the two-party dialogue you generate contains only the speaker1's utterance, not the speaker2's response. The multi-choice question you generated should be as difficult as possible. The content of the last turn of dialogue must be related to the whole dialogue, not only to the utterances of the most recent turns. All candidates must include the commonsense.  At the same time, all of the candidates seem to be plausible responses, but only one commonsense they contain is correct. This commonsense is what a sane adult living in society can judge between right and wrong. None of the candidate options can be literally similar or overlap with the generated commonsense.

~

The format of the question is as follows: \{"id": "xx", "task": "Commonsense-aware Dialogue Generation", "domain": "\{Domain\}", "speaker1 personality": "xx", "speaker2 personality": "xx",  "dialogue": [\{"speaker1": "xx", "speaker2": "xx"\}, \{"speaker1": "xx", "speaker2": "xx"\}, \{"speaker1": "xx", "speaker2": "xx"\}, ..., \{"speaker1": "xx"\}], "commonsense": "xx", "option": "xx", "analysis": "xx", "label": "xx"\}. Where "id" denotes a randomly generated id, "speaker1 personality" denotes the character personality of speaker1, "speaker2 personality" denotes the character personality of speaker2, "dialogue" denotes more than 10 turns of two-person English dialogue where the two speakers are represented by speaker1 and speaker2  in each turn of interaction, and the last turn contains only speaker1 words and cannot be a question. "commonsense" is the used commonsense, "option" denotes a dictionary that contains 4 candidate options, with capital letter sequence identifiers as the key. "analysis" denotes the reason for choosing that correct option. "label" denotes the correct option represented by a capital letter. Note that the generated question should be in JSON format. Now, please set the question. \\

    \bottomrule
\end{tabularx}}
\caption{The prompt for data generation of commonsense-aware response generation by GPT-4.}
\end{table*}

\begin{table*}[t]
\centering
\resizebox{\textwidth}{!}{
\begin{tabularx}{\textwidth}
{>{\raggedright\arraybackslash}p{16cm}}
    \toprule
    \textbf{Prompt for Knowledge-grounded Response Generation} \\
    \midrule
We are currently testing the annotation capabilities of data annotators for response selection in multi-turn knowledge-grounded dialogues. Multi-turn dialogue refers to a chat record generated by two speakers engaging in continuous interactions using natural language. The knowledge-grounded response selection means that according to the speaker1's response in the last turn of the multi-turn dialogue, the annotators must select the most correct and appropriate response from the candidates, considering the context of the multi-turn dialogue and the given background knowledge.
You are an expert at knowledge-grounded response selection in multi-turn dialogues, and now please act as the test creator. You first need to generate a paragraph of background knowledge of more than 1000 words, then generate more than 10 turns(10 turns=20 utterances) of two-party dialogue on the topic of background knowledge. Furthermore, you should write a multi-choice question that requires you to select the correct knowledge from the background knowledge to answer. The dialogue content and background knowledge are required to be relevant to \{Domain\}.

~

Before setting the question, you should randomly set the personality of both speakers, and you have a certain probability of setting the personality of the speakers to be unfriendly so as to generate unfriendly, ironic, offensive, quarreling, specifying, weird, and negative content. It simulates the dialogue scenes of real human beings and helps to better study the response selection ability in multi-turn knowledge-grounded dialogues under various scenarios.

~

When you generate the dialogue, please avoid using interrogative sentences in the dialogue. The last turn of the dialogue is not the end of the dialogue session. The last turn of the two-party dialogue you generate contains only the utterance of speaker1, not the response of speaker2. The response in the last turn can only be generated based on the dialogue and background knowledge, with no additional external knowledge required. The multi-choice question you generated should be as difficult as possible. All candidates must be connected to the dialogue, and the knowledge used must come from background knowledge. However, none of the candidate options can literally resemble or overlap the background knowledge.  In particular, the wrong option cannot be a negative expression of a piece of background knowledge. All candidate options require comprehensive understanding and extended reasoning dialogue, and background knowledge to determine the plausibility of the options. In addition, all candidate options must be similar enough in terms of length and content, making it more difficult to distinguish between right and wrong candidate options.

~

The format of the question is as follows: \{"id": "xx", "task": "Knowledge-grounded Response Generation", "domain": "\{Domain\}", "speaker1 personality": "xx",  "speaker2 personality": "xx",  "knowledge": "xx", "dialogue": [\{"speaker1": "xx", "speaker2": "xx"\}, \{"speaker1": "xx", "speaker2": "xx"\},\{"speaker1": "xx", "speaker2": "xx"\}, ..., \{"speaker1": "xx"\}],  "option": "xx", "analysis": "xx", "label": "xx"\}. Where "id" denotes a randomly generated id, "speaker1 personality" denotes the character personality of speaker1, "speaker2 personality" denotes the character personality of speaker2. "knowledge" is background knowledge of text longer than 1000 words. "dialogue" denotes more than 10 turns of two-party English dialogue where the two speakers are represented by speaker1 and speaker2  in each turn of interaction, and the last turn only contains the utterance of speaker1 and cannot be a question., "option" denotes a dictionary that contains 4 candidate options, with capital letter sequence identifiers as the key. "analysis" denotes the reason for choosing that correct option. "label" denotes the correct option represented by a capital letter. Note that the generated question should be in JSON format. Now, please set the question. \\

    \bottomrule
\end{tabularx}}
\caption{The prompt for data generation of knowledge-grounded response generation by GPT-4.}
\end{table*}

\begin{table*}[t]
\centering
\resizebox{\textwidth}{!}{
\begin{tabularx}{\textwidth}
{>{\raggedright\arraybackslash}p{16cm}}
    \toprule
    \textbf{Prompt for Dialogue Natural Language Inference}\\
    \midrule
We are currently testing the annotation capabilities of data annotators for dialogue natural language inference in multi-turn dialogues. Multi-turn dialogue refers to a chat record generated by two speakers engaging in continuous interactions using natural language. Natural language inference refers to giving an utterance of a speaker in a multi-turn dialogue as a premise and giving a hypothesis at the same time, and semantic analysis is carried out to determine the semantic relationship between the premise and the hypothesis, including entailment, contradiction, and neutral.
You are an expert at natural language inference in multi-turn dialogues, and now please act as the test creator. You first need to assume a semantic relationship of \{Relation\}. 

~

Then, you should generate a two-party dialogue with at least 10 turns(10 turns = 20 utterances) and specify an utterance of a speaker at a turn as the premise. At the same time, you should generate a hypothesis that fits the specified semantic relationship. Finally, write a multi-choice question of dialogue natural language inference that requires a comprehensive understanding of the context of multi-turn dialogue. The dialogue content should be relevant to 
Before setting the question, you should randomly set the personality of both speakers, and you have a certain probability of setting the personality of the speakers to be unfriendly so as to generate unfriendly, ironic, offensive, quarreling, specifying, weird, and negative content. It simulates the dialogue scenes of real human beings and helps to better study the dialogue natural language inference ability under various scenarios.

~

When you generate the dialogue, please avoid using interrogative sentences in the dialogue. The last turn of the dialogue is not the end of the dialogue session. The multi-choice question you generated should be as difficult as possible. When the semantic relation belongs to entailment, the hypothesis can be fully inferred from the premise. When the semantic relation belongs to contradiction, the premise can fully infer the negation of the hypothesis. When the semantic relation belongs to neutral, the premise neither entails nor contradicts the hypothesis, and the semantic relation between the premise and hypothesis belongs to other cases except entailment and contradiction.

~

The format of the question is as follows: \{"id": "xx", "task": "Dialogue NLI", "semantic relationship": "\{Relation\}",  "domain": "\{Domain\}", "speaker1 personality": "xx", "speaker2 personality": "xx",  "dialogue": [\{"speaker1": "xx", "speaker2": "xx"\}, \{"speaker1": "xx", "speaker2": "xx"\}, \{"speaker1": "xx", "speaker2": "xx"\}, ..., \{"speaker1": "xx", "speaker2": "xx"\}], "premise": "xx", "hypothesis": "xx", "option": "xx", "analysis": "xx", "label": "xx" \}. Where "id" denotes a randomly generated id, "speaker1 personality" denotes the character personality of speaker1, "speaker2 personality" denotes the character personality of speaker2, "dialogue" denotes more than 10 turns of two-person English dialogue where the two speakers are represented by speaker1 and speaker2  in each turn of interaction. "premise" is a specified premise, and "hypothesis" is a hypothesis that conforms to a given semantic relationship with the premise, "option" denotes a dictionary that contains 3 candidate options, with capital letter sequence identifiers as the key, the candidate options are fixed and include three types: entailment, contradiction and neutrality, of which only one type is correct. "analysis" denotes the reason for choosing that correct option. "label" denotes the correct option represented by a capital letter. Note that the generated question should be in JSON format. Now, please set the question.\\

    \bottomrule
\end{tabularx}}
\caption{The prompt for data generation of dialogue natural language inference by GPT-4.}
\end{table*}

\begin{table*}[t]
\centering
\resizebox{\textwidth}{!}{
\begin{tabularx}{\textwidth}
{>{\raggedright\arraybackslash}p{15.5cm}}
    \toprule
    \textbf{Prompt for Offensive Detection}\\
    \midrule
We are currently testing the annotation capabilities of data annotators for offensive detection in multi-turn dialogues. Multi-turn dialogue refers to a chat record generated by two speakers engaging in continuous interactions using natural language. Offensive detection refers to judging whether utterances of the speaker contain uncivil, discriminatory, or aggressive content in a multi-turn dialogue.

~

You are an expert at offensive detection in multi-turn dialogues, and now please act as the test creator. You first limit the dialogue generated to \{contain or not contain\} offensive remarks. Then, you should generate a two-party dialogue with at least 10 turns(10 turns = 20 utterances) that meets the above requirements. Finally, you need to generate a multi-choice question about offensive detection in the multi-turn dialogue. The dialogue content should be relevant to \{Domain\}.

~

Before setting the question, you should randomly set the personality of both speakers, and you have a certain probability of setting the personality of the speakers to be unfriendly so as to generate unfriendly, ironic, offensive, quarreling, specifying, weird, and negative content. It simulates the dialogue scenes of real human beings and helps to better study the offensive detection ability under various scenarios.

~

When you generate the dialogue, please avoid using interrogative sentences in the dialogue.  The last turn of the dialogue is not the end of the dialogue session.  The multi-choice question you generated should be as difficult as possible.  The content of the dialogue may contain some offensive statements.  If the dialogue you generate contains offensive statements, one turn will be offensive at most, helping make your question difficult. The offensive statement you generate cannot be literally offensive in order to increase the difficulty of the test.

~

The format of the question is as follows: \{"id": "xx", "task": "Offensive Detection", 
"offensive": "\{contain or not contain\}", "domain": "\{Domain\}",  "speaker1 personality": "xx", "speaker2 personality": "xx", "dialogue": [\{"speaker1": "xx", "speaker2": "xx"\}, \{"speaker1": "xx", "speaker2": "xx"\}, \{"speaker1": "xx", "speaker2": "xx"\}, ..., \{"speaker1": "xx", "speaker2": "xx"\}], "option": "xx", "analysis": "xx", "label": "xx"\}. Where "id" denotes a randomly generated id, "speaker1 personality" denotes the character personality of speaker1, "speaker2 personality" denotes the character personality of speaker2, "dialogue" denotes more than 10 turns of two-person English dialogue where the two speakers are represented by speaker1 and speaker2  in each turn of interaction. "option" denotes a dictionary that contains 2 candidate options, with capital letter sequence identifiers as the key, the candidate options are fixed and include two types: offensive and non-offensive, of which only one type is correct. "analysis" denotes the reason for choosing that correct option. "label" denotes the correct option represented by a capital letter. Note that the generated question should be in JSON format. Now, please set the question.\\

    \bottomrule
\end{tabularx}}
\caption{The prompt for data generation of offensive detection by GPT-4.}
\end{table*}

\begin{table*}[t]
\centering
\resizebox{\textwidth}{!}{
\begin{tabularx}{\textwidth}
{>{\raggedright\arraybackslash}p{15.5cm}}
    \toprule
    \textbf{Prompt for Personality-grounded Response Generation} \\
    \midrule
We are testing data annotators on their annotation capabilities for response selection in multi-turn personality-grounded dialogues. A multi-turn dialogue is a chat record generated by multiple turns of continuous interaction between two speakers using natural language. The persona-grounded response selection refers to the fact that for the last turn of the multi-turn persona-grounded dialogues, the annotator needs to combine the multi-turn dialogues context and the given speaker2's personality to select the most appropriate response from the candidate responses that reflects the given persona-grounded dialogues. 

~

You are an expert at persona-grounded response selection in multi-turn dialogues, and now please act as the test creator. You first need to generate a detailed personality of speaker2, followed by a two-person dialogue with more than 10 turns, where the last turn of the two-person multi-turn dialogue contains only speaker1's words and not speaker2's response. Then a response selection multi-choice question in multi-turn persona-grounded dialogues is also generated, where each candidate option is a candidate response from speaker2 in response to the dialogues above. The dialogue is required to be related to \{Domain\}. 

~

Before setting the question, you should randomly set the personality of both speakers, and you have a certain probability of setting the personality of the speakers to be unfriendly so as to generate unfriendly, ironic, offensive, quarreling, specifying, weird, and negative content. It simulates the dialogue scenes of real human beings and helps to better study the intent understanding ability under the various scenarios.
When you generate the dialogue, please avoid using interrogative sentences in the dialogue.  The content of the last turn of the sub-dialogue is not a closing statement. The questions you come up with must be as difficult as possible, and the content of the last turn's dialogue must be coherent with the entire dialogue, not just the sentences from the most recent turn. At the same time, the words of speaker1 in the final turn must be able to detect the personality of the other speaker in a targeted way. For this reason, the content of the last turn cannot be the closing sentence of the entire dialogue. None of the candidate options can literally resemble or overlap with that personality. In addition, the candidate choices are all coherent and reasonable responses to the above part of the dialogue, and they differ only in personality, which can be clearly distinguished after a comprehensive understanding of the entire dialogue and the personality. The choice of which option is the correct answer is also based only on the degree of personality.

~

The format of the question is as follows: \{"id": "xx", "task": "Personality-grounded Response Generation",  "domain": "\{Domain\}", "speaker1 personality": "xx",  "speaker2 personality": "xx", "persona": "xx",  "dialogue": [\{"speaker1": "xx", "speaker2": "xx"\}, \{"speaker1": "xx", "speaker2": "xx"\}, \{"speaker1": "xx", "speaker2": "xx"\}, ..., \{"speaker1": "xx"\}], "option": "xx", "analysis": "xx", "label": "xx"\}. Where "id" denotes a randomly generated id, "speaker1 personality" denotes the character personality of speaker1, "speaker2 personality" denotes the character personality of speaker2. "persona" is a description of the speaker2's personality in English, "dialogue" denotes more than 10 turns of two-party English dialogue where the two speakers are represented by speaker1 and speaker2  in each turn of interaction, and the last turn only contains the utterance of speaker1 and cannot be a question, "option" denotes a dictionary that contains 4 candidate options, with capital letter sequence identifiers as the key. "analysis" denotes the reason for choosing that correct option. "label" denotes the correct option represented by a capital letter. Note that the generated question should be in JSON format. Now, please set the question. \\

    \bottomrule
\end{tabularx}}
\caption{The prompt for data generation of Personality-grounded Dialogue Generation by GPT-4.}
\end{table*}

\begin{table*}[t]
\centering
\resizebox{\textwidth}{!}{
\begin{tabularx}{\textwidth}
{>{\raggedright\arraybackslash}p{16cm}}
    \toprule
    \textbf{Prompt for Dialogue Infilling} \\
    \midrule
We are currently testing the annotation capabilities of data annotators for dialogue infilling in multi-turn dialogues. Multi-turn dialogue refers to a chat record generated by two speakers engaging in continuous interactions using natural language. Dialogue infilling means, in the last turn of a multi-turn dialogue, the utterance spoken by speaker1 is unknown, and the response spoken by speaker2 is known.  Then, the annotator needs to predict what utterance spoken by the speaker1 should be in the last turn in consideration of the context of the multi-turn dialogue and the response spoken by the speaker2.
You are an expert at dialogue infilling in multi-turn dialogues, and now please act as the test creator. Firstly, please generate a two-party dialogue with at least 10 turns(10 turns = 20 utterances). Then, you generate 4 candidate utterances spoken by the speaker1 based on the dialogue content. Finally, randomly select one of the candidate utterances of speaker1 to generate a response from speaker2 who can answer the utterance. The dialogue content should be relevant to \{Domain\}.

Before setting the question, you should randomly set the personality of both speakers, and you have a certain probability of setting the personality of the speakers to be unfriendly so as to generate unfriendly, ironic, offensive, quarreling, specifying, weird, and negative content. It stimulates the dialogue scenes of real human beings, and helps to better study the dialogue-infilling ability under the various scenarios.

When you generate the dialogue, please avoid using interrogative sentences in the dialogue. The last turn of the dialogue is not the end of the dialogue session. The multi-choice question you generated should be as difficult as possible. The utterance that you generated spoken by speaker1 must be related to the whole dialogue so that the annotators must rely more on the dialogue context to make decisions. In addition, each candidate of the multi-choice question is as similar as possible, but there are good and bad differences between the options, and these differences can be distinguished by a comprehensive understanding of the dialogue above and the response spoken by speaker2 in the last turn. Finally, there should be no literal similarity or overlap between speaker1's utterance and speaker2's responses that you generate, lest the annotator filter the answers directly by literal similarity.

The format of the question is as follows:  \{"id": "xx", "task": "Dialogue Infilling", "domain": "\{Domain\}", "speaker1 personality": "xx", "speaker2 personality": "xx", "dialogue": [\{"speaker1": "xx", "speaker2": "xx"\}, \{"speaker1": "xx", "speaker2": "xx"\}, \{"speaker1": "xx", "speaker2": "xx"\}, ..., \{"speaker1": "xx", "speaker2": "xx"\}], "option": "xx", "analysis": "xx", "label": "xx", "response": "xx"\}. Where "id" denotes a randomly generated id, "speaker1 personality" denotes the character personality of speaker1, "speaker2 personality" denotes the character personality of speaker2, "dialogue" denotes more than 10 turns of two-person English dialogue where the two speakers are represented by speaker1 and speaker2  in each turn of interaction, it does not include the perosn1's utterance to be completed and the following speaker2's response. "option" denotes a dictionary that contains 4 candidate options, with capital letter sequence identifiers as the key. Each of these options is a possible candidate for speaker1's utterance. "analysis" denotes the reason for choosing that correct option. "label" denotes the correct option represented by a capital letter. Note that the generated question should be in JSON format. Now, please set the question. \\

    \bottomrule
\end{tabularx}}
\caption{The prompt for data generation of dialogue infilling by GPT-4.}
\end{table*}

\begin{table*}[t]
\centering
\resizebox{\textwidth}{!}{
\begin{tabularx}{\textwidth}
{>{\raggedright\arraybackslash}p{16cm}}
    \toprule
    \textbf{Prompt for Relation Classification} \\
    \midrule
We are currently testing the annotation capabilities of data annotators for relation classification in multi-turn dialogues. Multi-turn dialogue refers to a chat record generated by two speakers engaging in continuous interactions using natural language. Relation classification means predicting the relationship type between the two dialogue speakers based on the context of the multi-turn dialogue.

~

You are an expert at relation classification in multi-turn dialogues, and now please act as the test creator. Firstly, you should assume the relationship between the two sides of the dialogue is \{Relation\}. Then, you generate a two-party dialogue with at least 10 turns(10 turns = 20 utterances) that fits the relationship. It is followed by a multi-choice question of relation classification that requires a comprehensive understanding of the context to answer. The dialogue content should be relevant to \{Domain\}.

~

Before setting the question, you should randomly set the personality of both speakers, and you have a certain probability of setting the personality of the speakers to be unfriendly so as to generate unfriendly, ironic, offensive, quarreling, specifying, weird, and negative content. It simulates the dialogue scenes of real human beings, and helps to better study the relation classification ability under the various scenarios.
When you generate the dialogue, please avoid using interrogative sentences in the dialogue.  The last turn of the dialogue is not the end of the dialogue session.  The multi-choice question you generated should be as difficult as possible.  The relationship between two speakers cannot be directly reflected in the dialogue content, but the relationship can be clearly judged by understanding the dialogue content and the implicit information contained therein, such as the dialogue tone, attitude, scene, and other rich information.  Candidates need to be as similar as possible, which is more confusing, but with a deep understanding of the context of the dialogue, the annotators can pick the right answer without debate.

~

The format of the question is as follows: \{"id": "xx", "task": "Relation Classification", "relation": "\{Relation\}", "domain": "\{Domain\}", "speaker1 personality": "xx",  "speaker2 personality": "xx",  ‘’dialogue": [\{"speaker1": "xx", "speaker2": "xx"\}, \{"speaker1": "xx", "speaker2": "xx"\}, \{"speaker1": "xx", "speaker2": "xx"\}, ..., \{"speaker1": "xx", "speaker2": "xx"\}], "option": "xx", "analysis": "xx", "label": "xx" \}. Where "id" denotes a randomly generated id, "relation" is the relation between two parties, "speaker1 personality" denotes the character personality of speaker1, "speaker2 personality" denotes the character personality of speaker2, "dialogue" denotes a more than 10 turns of two-person English dialogue where the two speakers are represented by speaker1 and speaker2  in each turn of interaction. "option" is a dictionary containing four candidate options, with capital letters representing the serial number identifier as the key. The candidate emotions are limited to some types: teacher-student, child-parent, child-other family elder, brothers, sisters, couples, lovers, friends, neighbors, colleagues, superior-subordinate, customers, competitors, of which only one type is correct. "analysis" denotes the reason for choosing that correct option. "label" denotes the correct option represented by a capital letter. Note that the generated question should be in JSON format. Now, please set the question. \\
    \bottomrule
\end{tabularx}}
\caption{The prompt for data generation of relation classification by GPT-4.}
\end{table*}

\begin{table*}[t]
\centering
\resizebox{\textwidth}{!}{
\begin{tabularx}{\textwidth}
{>{\raggedright\arraybackslash}p{15.5cm}}
    \toprule
    \textbf{Prompt for Multi-turn Response Generation} \\
    \midrule
We are currently testing the annotation capabilities of data annotators for response selection in multi-turn dialogues. Multi-turn dialogue refers to a chat record generated by two speakers engaging in continuous interactions using natural language. Response selection refers to the process in which the annotators select the correct response from the candidates as the most coherent and reasonable response in the dialogue after comprehensively understanding the content of the multi-turn dialogue.

~

You are an expert at response selection in multi-turn dialogues, and now please act as the test creator. You first need to generate more than 10 turns(10 turns=20 utterances) of two-party dialogue. Then, you need to generate a multi-turn dialogue response selection question that requires a comprehensive understanding of the dialogue context. The dialogue content is required to be relevant to \{Domain\}.
Before setting the question, you should randomly set the personality of both speakers, and you have a certain probability of setting the personality of the speakers to be unfriendly so as to generate unfriendly, ironic, offensive, quarreling, specifying, weird, and negative content. It simulates the dialogue scenes of real human beings and helps to better study the response selection ability in multi-turn dialogues under various scenarios.

~

When you generate the dialogue, please avoid using interrogative sentences in the dialogue. The last turn of the dialogue is not the end of the dialogue session. Each candidate of the question matches the personality of speaker2. When analyzing which choice is the correct answer, do not mention the speaker2's personality. In addition, each candidate of the question is only good or bad in terms of coherence. That is, the most coherent response is an effective continuation of the conversation above, which is an orderly chain of events under a common theme with the topic discussed above. The wrong choice may refer to a topic far removed from the one discussed in the conversation. These differences can be easily distinguished after a comprehensive understanding of the entire dialogue so that the annotator can pick the correct choice without dispute. Finally, in order to ensure fairness, the generated dialogue and options do not involve external specific knowledge as much as possible. If external specific knowledge must be involved, try to be common sense as much as possible, and do not involve unfamiliar concepts or entities.

~

The format of the question is as follows: \{"id": "xx", "task": "Multi-turn Response Generation", "domain": "\{Domain\}", "speaker1 personality": "xx",  "speaker2 personality": "xx", "dialogue": [\{"speaker1": "xx", "speaker2": "xx"\}, \{"speaker1": "xx", "speaker2": "xx"\}, \{"speaker1": "xx", "speaker2": "xx"\}, ..., \{"speaker1": "xx"\}], "option": "xx", "analysis": "xx", "label": "xx"\}. Where "id" denotes a randomly generated id, "speaker1 personality" denotes the character personality of speaker1, "speaker2 personality" denotes the character personality of speaker2. "dialogue" denotes more than 10 turns of two-party English dialogue where the two speakers are represented by speaker1 and speaker2  in each turn of interaction, and the last turn only contains the utterance of speaker1 and cannot be a question, "option" denotes a dictionary that contains 4 candidate options, with capital letter sequence identifiers as the key. "analysis" denotes the reason for choosing that correct option. "label" denotes the correct option represented by a capital letter. Note that the generated question should be in JSON format. Now, please set the question. \\

    \bottomrule
\end{tabularx}}
\caption{The prompt for data generation of multi-turn response generation by GPT-4.}
\end{table*}

\begin{table*}[t]
\centering
\resizebox{\textwidth}{!}{
\begin{tabularx}{\textwidth}
{>{\raggedright\arraybackslash}p{15.5cm}}
    \toprule
    \textbf{Prompt for Slot Filling}\\
    \midrule
We are currently testing the annotation capabilities of data annotators for multi-turn dialogue slot filling. Multi-turn dialogue refers to chat records generated by continuous interaction between two speakers using natural language. Slot filling refers to extracting the corresponding value for a specific slot (such as time, location, name, etc.) from the events reflected in multi-turn dialogues. 

~

You are an expert in multi-turn dialogue slot filling tasks, and now you are asked to be the question setter. First, you assume the slot of interest, then generate a more than 10-turn dialogue between two people that involves the slot multiple times. After that, create a multi-turn dialogue slot filling multi-choice question that requires a comprehensive understanding of the dialogue context to answer. The dialogue content must be related to \{Domain\}. 

~

Before setting the questions, you need to randomly set the personalities of the two speakers. There is a certain probability that you will set the speaker's personality to be unfriendly, generating unfriendly, sarcastic, offensive, argumentative, sophistical, weird, or negative content. This simulates real human dialogue scenarios and helps to better study slot filling capabilities in rich scenarios. 

~

Try not to use questions in the generated multi-turn dialogue. The last turn of dialogue should not be a closing statement. Your question must be as difficult as possible, requiring a comprehensive understanding of the entire dialogue, rather than focusing on a single turn or sentence. At the same time, the information involved in all candidate options must come from the dialogue, and all candidate options are seemingly correct values for the given slot. So, you need to make sure that these candidate values appear when generating multi-turn dialogues. The question must be answerable based on the multi-turn dialogue and does not require external knowledge. In particular, all candidate options should be similar enough in length and content to make it more difficult to distinguish between correct and incorrect options. All candidate options need to comprehensively understand and extend the reasoning of the multi-turn dialogue content to judge the correctness of the options. 

~

The format of the question is as follows:\{"id": "xx", "task": "Slot Filling", "slot": "xx", "domain": "\{Domain\}", "speaker1 personality": "xx", "speaker2 personality": "xx", "dialogue": [\{"speaker1": "xx", "speaker2": "xx"\}, \{"speaker1": "xx", "speaker2": "xx"\}, \{"speaker1": "xx", "speaker2": "xx"\}, ..., \{"speaker1": "xx", "speaker2": "xx"\}], "test\_question": "xx", "option": "xx", "analysis": "xx", "label": "xx"\}. Where "id" is a randomly generated id, "slot" is the focused slot, "speaker1 personality" is the personality of speaker1, "speaker2 personality" is the personality of speaker2, "dialogue" is a more than 10-turn English dialogue between two people, with each turn of interaction represented by speaker1 and speaker2. "test\_question" is the multi-choice question, "option" denotes a dictionary that contains 4 candidate options, with capital letter sequence identifiers as the key. "analysis" denotes the reason for choosing that correct option. "analysis" is the reason for choosing the option. "label" is the correct option corresponding to the order. Please note that your generated test questions must be in JSON format. Now please set the questions. \\
    \bottomrule
\end{tabularx}}
\caption{The prompt for data generation of slot filling by GPT-4.}
\end{table*}

\begin{table*}[t]
\centering
\resizebox{\textwidth}{!}{
\begin{tabularx}{\textwidth}
{>{\raggedright\arraybackslash}p{15.5cm}}
    \toprule
    \textbf{Prompt for Dialogue Summarization} \\
    \midrule
We are testing data annotators on their ability to annotate multi-turn dialogue summarizations. A multi-turn dialogue is a chat transcript produced by multiple turns of continuous interaction between two speakers using natural language. Dialogue summarization is the process of extracting, summarizing, or refining key information from a multi-turn dialogue, turning it into a short summary paragraph that can be used to present the main points or big ideas of that multi-turn dialogue. 

~

You are an expert at dialogue summarization in multi-turn dialogues, and now please act as the test creator. Firstly, you should generate a two-party dialogue with at least 10 turns(10 turns = 20 utterances). Then, you should create a dialogue summarization multiple-choice question that requires a comprehensive understanding of the context of the multi-turn dialogue. The dialogue content should be relevant to \{Domain\}.

~

Before setting the question, you should randomly set the personality of both speakers, and you have a certain probability of setting the personality of the speakers to be unfriendly so as to generate unfriendly, ironic, offensive, quarreling, specifying, weird, and negative content. It simulates the dialogue scenes of real human beings and helps to better study the intent understanding ability under the various scenarios.

~

The question you come up with must be as difficult as possible, but there are clear differences in strengths and weaknesses between each of the candidate options for multi-choice questions in terms of (1) inconsistencies between the information in the summarization and in the given dialogue, (2) the presence of information in the summarization that is not present in the given dialogue, and (3) the loss of important information in the given dialogue in the summarization, which can be clearly differentiated by a synthesized comprehension of the entire dialogue content. Candidate options only generate summarizations of the dialogue content, and it is not necessary to refer to the character names and personality information of the two speakers when judging which option to choose as the correct answer, but only from the degree of comprehensiveness of the summarization in presenting the main content of the dialogue.

~

The format of the question is as follows: \{"id": "xx", "task": "Dialogue Summarization", "domain": "\{Domain\}", "speaker1 personality": "xx", "speaker2 personality": "xx", "dialogue": [\{"speaker1": "xx", "speaker2": "xx"\}, \{"speaker1": "xx", "speaker2": "xx"\}, \{"speaker1": "xx", "speaker2": "xx"\}, ..., \{"speaker1": "xx", "speaker2": "xx"\}],"option": "xx", "analysis": "xx", "label": "xx"\}. Where "id" denotes a randomly generated id, "speaker1 personality" denotes the character personality of speaker1, "speaker2 personality" denotes the character personality of speaker2, "dialogue" denotes more than 10 turns of two-person English dialogue where the two speakers are represented by speaker1 and speaker2  in each turn of interaction. "option" denotes a dictionary that contains 4 candidate options, with capital letter sequence identifiers as the key. "analysis" denotes the reason for choosing that correct option. "label" denotes the correct option represented by a capital letter. Note that the generated question should be in JSON format. Now, please set the question. \\

    \bottomrule
\end{tabularx}}
\caption{The prompt for data generation of dialogue summarization by GPT-4.}
\label{tab:ds_prompt}
\end{table*}

\begin{table*}[t]
\centering
\resizebox{\textwidth}{!}{
\begin{tabularx}{\textwidth}
{>{\raggedright\arraybackslash}p{15.5cm}}
    \toprule
    \textbf{Prompt for Data Filter} \\
    \midrule
 As a data quality inspector, you should conduct a comprehensive quality assessment of the following multi-choice question related to \{task\}. Your assessment needs to take into account both the correctness of the test question and the answer. 
 Specifically, question correctness refers to whether the question is clear and relevant to the given dialogue and external knowledge if exists.
 Answer correctness refers to whether the content corresponding to the given label can correctly answer the given question.
 Next, please assess whether the given multi-choice question is correct. Do not analyze and directly give "correct" or "incorrect" as the output.
 
~

[Dialogue]

\{dialogue\_content\}

~

[Test Question]

\{test\_question\}

~

[Options]

\{option\_contents\}

~

[Answer]

\{answer\_label\}

 \\

    \bottomrule
\end{tabularx}}
\caption{The prompt for data filtering.}
\label{tab:filter_prompt}
\end{table*}

\end{document}